\documentclass[twocolumn]{article}

\usepackage{PRIMEarxiv}

\usepackage[utf8]{inputenc} 
\usepackage[T1]{fontenc}    
\usepackage{hyperref}       
\usepackage{url}            

\usepackage{amsmath}
\usepackage{amsfonts}
\usepackage{amssymb}

\usepackage{booktabs}
\usepackage{multirow}
\usepackage{array}

\usepackage[table,dvipsnames]{xcolor}
\definecolor{softgreen}{RGB}{200,230,201}

\usepackage{graphicx}
\graphicspath{{images/}{media/}}
\usepackage{caption}
\usepackage{subcaption}

\usepackage{tikz}
\usepackage{pgfplots}
\pgfplotsset{compat=1.18}
\usetikzlibrary{calc}
\pgfplotsset{
  YOLOv11n/.style={color=blue},
  YOLOv11s/.style={color=orange},
  YOLOv11m/.style={color=green!60!black},
  fp32/.style={solid, mark=o},
  int8/.style={dashed, mark=triangle*},
}

\usepackage{verbatim}
\usepackage{nicefrac}
\usepackage{microtype}
\usepackage{fancyhdr}

\title{Fully-sensorized smart-eyewear platform for on-device Machine Learning
\thanks{This work was carried out in the EssilorLuxottica "Smart Eyewear Lab", a Joint Research Center between EssilorLuxottica and Politecnico di Milano.}
}

\author{
  Andrea Giudici, Christian Veronesi, Pietro Bartoli, Mario Caliò, Franco Zappa \\
  DEIB, Politecnico di Milano \\
  Milano, Italy \\
  \texttt{name.surname@polimi.it} \\
  \And
  Aurelio Teliti, Giacomo Gervasoni, Diana Trojaniello \\
  EssilorLuxottica \\
  Milano, Italy \\
  \texttt{name.surname@essilorluxottica.com} \\
}

\begin{document}

\makeatletter
\twocolumn[
  \begin{@twocolumnfalse}
    \maketitle
    \begin{abstract}
    This paper presents ARGO, a smart eyewear platform designed to bridge ergonomic comfort, high computational throughput, and energy efficiency. Unlike cloud-dependent solutions, ARGO leverages the STM32N6 microcontroller and its integrated Neural Processing Unit (NPU) to enable on-device machine learning, minimizing latency and preserving user privacy through local data processing.
    The primary contribution lies in the holistic co-design of hardware, firmware, and artificial intelligence, centered on the deployment of an optimized YOLOv11 model for real-time urban obstacle recognition.
    To ensure compatibility with the target NPU, we introduce Head-wise Parallel Attention (HPA), an architectural refinement that enables efficient accelerator execution while preserving the original computational logic.
    The model is trained on the Walking On The Road (WOTR) dataset, and the final deployed configuration achieves an \(mAP_{50\text{--}95}\) of 24\% under strict memory constraints, with a memory footprint of only 2.483 MB.
    The platform integrates a multimodal sensor suite --- RGB cameras, Time-of-Flight sensors, microphones, and ambient sensors --- and delivers 10 FPS at a continuous autonomy of ~113 minutes on a 200 mAh battery. These results demonstrate the feasibility of a high-performance, privacy-preserving, and socially acceptable assistive device, and highlight how competitive edge AI solutions increasingly demand tightly integrated, multidisciplinary co-design approaches.
    \end{abstract}
    \vspace{0.4em}
    \vspace{1.2em}
  \end{@twocolumnfalse}
]
\makeatother

\section{Introduction}
\label{sec:introduction}

In recent years, interest in wearable technology and Internet of Things (IoT) has accelerated, driven not only by advances in ultra-low-power electronics and the progressive reduction of manufacturing costs, but also by improved wireless communication protocols, the integration of edge-AI capabilities, and the growing demand for continuous monitoring in consumer, industrial and healthcare applications~\cite{8736011, 8114708, bulling2014tutorial}.

In particular, wearable technologies have emerged as a key player in the transition toward ubiquitous healthcare and enhanced situational awareness. These devices facilitate the continuous, non-invasive acquisition and monitoring of critical physiological and biological parameters---including heart rate variability~\cite{alugubelli2022wearable} (HRV), blood glucose concentrations~\cite{cappon2017wearable}, and respiratory rate~\cite{angelucci2023imu}---throughout the user's daily activities. By providing real-time data streams, wearables offer invaluable insights for both clinical diagnostics and athletic performance optimization~\cite{11192188, s26031079}.

Simultaneously, the evolution of wearable systems extends beyond health monitoring to encompass the digital processing of the user's immediate environment.
Through the integration of Computer Vision (CV) algorithms, proximity sensors, and cameras, these devices can detect obstacles, identify potential hazards, and recognize specific objects within the user's vicinity. This dual functionality---combining internal biological data with external environmental mapping---paves the way for advanced assistive technologies and proactive safety systems in industrial and healthcare settings.

\begin{figure}[t]
\centering
\begin{tikzpicture}
\begin{axis}[
    width=\linewidth,
    height=0.62\linewidth,
    xlabel={\textbf{Year}},
    ylabel={\textbf{Publications}},
    xmin=1969, xmax=2024,
    ymode=log,
    log basis y=10,
    log ticks with fixed point,
    ymin=1, ymax=30000,
    ytick={1,10,100,1000,10000},
    xtick={1970,1990,2010,2020},
    xticklabel style={font=\footnotesize, anchor=north},
    yticklabel style={
        font=\footnotesize,
        /pgf/number format/1000 sep={}
    },
    xlabel style={font=\footnotesize},
    ylabel style={font=\footnotesize},
    grid=both,
    major grid style={line width=.2pt, draw=gray!40},
    minor grid style={line width=.1pt, draw=gray!20},
    tick style={black},
    enlargelimits=0.02,
    clip=false,
    scaled x ticks=false,
    x tick label style={
      /pgf/number format/1000 sep={}
    },
    axis line style={line width=0.6pt},
    tick align=outside,
    major tick length=2.2pt,
    minor tick length=1.2pt,
    ybar,
    bar shift=0pt,
    legend style={
        at={(0.5,-0.28)},
        anchor=north,
        font=\footnotesize,
        draw=none,
        fill=none,
        legend columns=2
    },
]
\addplot[
    draw=gray!60!black,
    fill=gray!60,
    fill opacity=0.25,
    bar width=2.6pt
] coordinates {
    (1969,72)(1970,78)(1971,83)(1972,63)(1973,117)(1974,124)(1975,152)(1976,136)
    (1977,153)(1978,148)(1979,167)(1980,185)(1981,218)(1982,193)(1983,194)(1984,223)
    (1985,234)(1986,237)(1987,212)(1988,236)(1989,273)(1990,269)(1991,281)(1992,316)
    (1993,327)(1994,331)(1995,403)(1996,532)(1997,578)(1998,638)(1999,670)(2000,712)
    (2001,715)(2002,851)(2003,978)(2004,1069)(2005,1317)(2006,1355)(2007,1654)(2008,1754)
    (2009,1996)(2010,2209)(2011,2530)(2012,2652)(2013,3195)(2014,4059)(2015,5825)(2016,7177)
    (2017,9005)(2018,10325)(2019,11851)(2020,12832)(2021,13923)(2022,15713)(2023,17120)(2024,19582)
};
\addlegendentry{Wearables}
\addplot[
    draw=blue!70!black,
    fill=blue!70!black,
    bar width=1.4pt
] coordinates {
    (1969,1)(1975,1)(1980,1)(1981,2)(1982,1)(1983,1)(1985,1)(1986,3)
    (1987,3)(1988,3)(1989,4)(1990,1)(1991,1)(1992,2)(1993,6)(1994,11)
    (1995,8)(1996,21)(1997,14)(1998,21)(1999,21)(2000,41)(2001,23)(2002,32)
    (2003,37)(2004,35)(2005,39)(2006,46)(2007,57)(2008,59)(2009,42)(2010,45)
    (2011,66)(2012,55)(2013,111)(2014,165)(2015,259)(2016,343)(2017,361)(2018,436)
    (2019,460)(2020,478)(2021,549)(2022,567)(2023,597)(2024,696)
};
\addlegendentry{Smart eyewear}
\end{axis}
\end{tikzpicture}
\caption{Annual publications for a macro \textit{wearables} query and its \textit{smart eyewear} subset. Data retrieved from Scopus (1969--2024).}
\label{fig:ew_trend}
\end{figure}
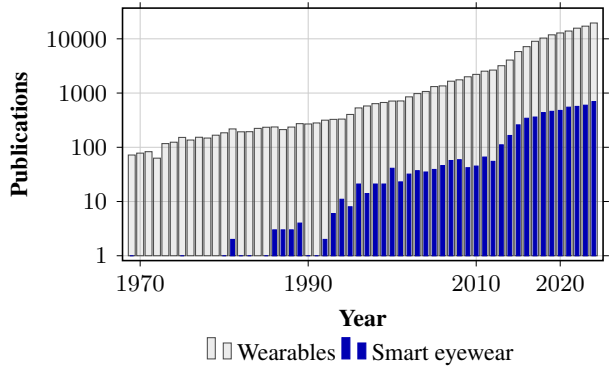

Within this context, smart eyewear has resurfaced as a prominent trend, as shown in Fig.~\ref{fig:ew_trend}, which highlights the trend of annual publications related to smart eyewear. These devices are electronic systems integrated into eyewear, capable of digitally monitoring and augmenting the wearer's analog perceptions and sensations through sensors.
While futuristic use-cases are compelling, practical deployments impose strict physical and computational constraints.
Eyewear that aims to be comfortable must necessarily be small, lightweight, and stylish, as they are worn on the face, the central point of human expression. This is even more true for those glasses worn every day, where the primary design paradigm is comfort and elegance. For this technology to be appreciated and used, it must first be comfortable and socially acceptable~\cite{rauschnabel2016augmented}.

In addition, the anatomical positioning of eyewear offers unparalleled advantages. The proximity to the primary sensory organs---specifically the eyes and ears---combined with a consistent egocentric perspective, enables the implementation of sophisticated algorithms for novel applications~\cite{7055926}. These include high-precision eye tracking for cognitive load estimation, ego-referenced gesture recognition for machine interaction, and attention monitoring~\cite{dell2025your, wang2025egoevgesture, 11105177, 11340225}. Furthermore, this unique viewpoint facilitates seamless Augmented Reality (AR) overlays and intuitive User Interface (UI) interactions that harmonize the digital layer with the user's physical surroundings~\cite{liu2022real}.

To achieve these objectives, the smart eyewear must be extensively sensorized to capture high-fidelity data from the surrounding environment. In this architecture, raw data undergoes an initial pre-processing stage directly on the device to reduce data dimensionality and extract relevant features. While these initial operations are computationally light, the subsequent extraction of high-level semantic information poses a significant challenge for the limited power budget of wearable hardware. However, resource-intensive computational tasks---specifically the execution of Deep Neural Networks (DNNs)---are managed through a tiered processing approach: localized execution on the dedicated hardware accelerator or, for more demanding tasks, offloading to external edge units or cloud services~\cite{electronics14071345}.

This hierarchical processing strategy is necessitated by the stringent power budget inherent to wearable platforms, particularly smart eyewear. To maintain an aesthetically pleasing and ergonomic industrial design, the physical volume allocated for power storage and electronics is severely constrained, typically residing within a volume of less than $5$ cm$^3$. This spatial limitation forces a critical trade-off between the form factor, battery capacity, system complexity, i.e. number of sensors, power architecture, memory space, and the computational complexity of the integrated sensor suite.

In this paper, we present ARGO, a smart eyewear system engineered to optimize the nexus between ergonomic comfort, high-performance computing, and energy efficiency. The core of the system architecture is the STM32N6 microcontroller (MCU), which provides a sophisticated equilibrium between peripheral integration, and CPU throughput.

\begin{figure}[t]
\centering
\includegraphics[width=\linewidth]{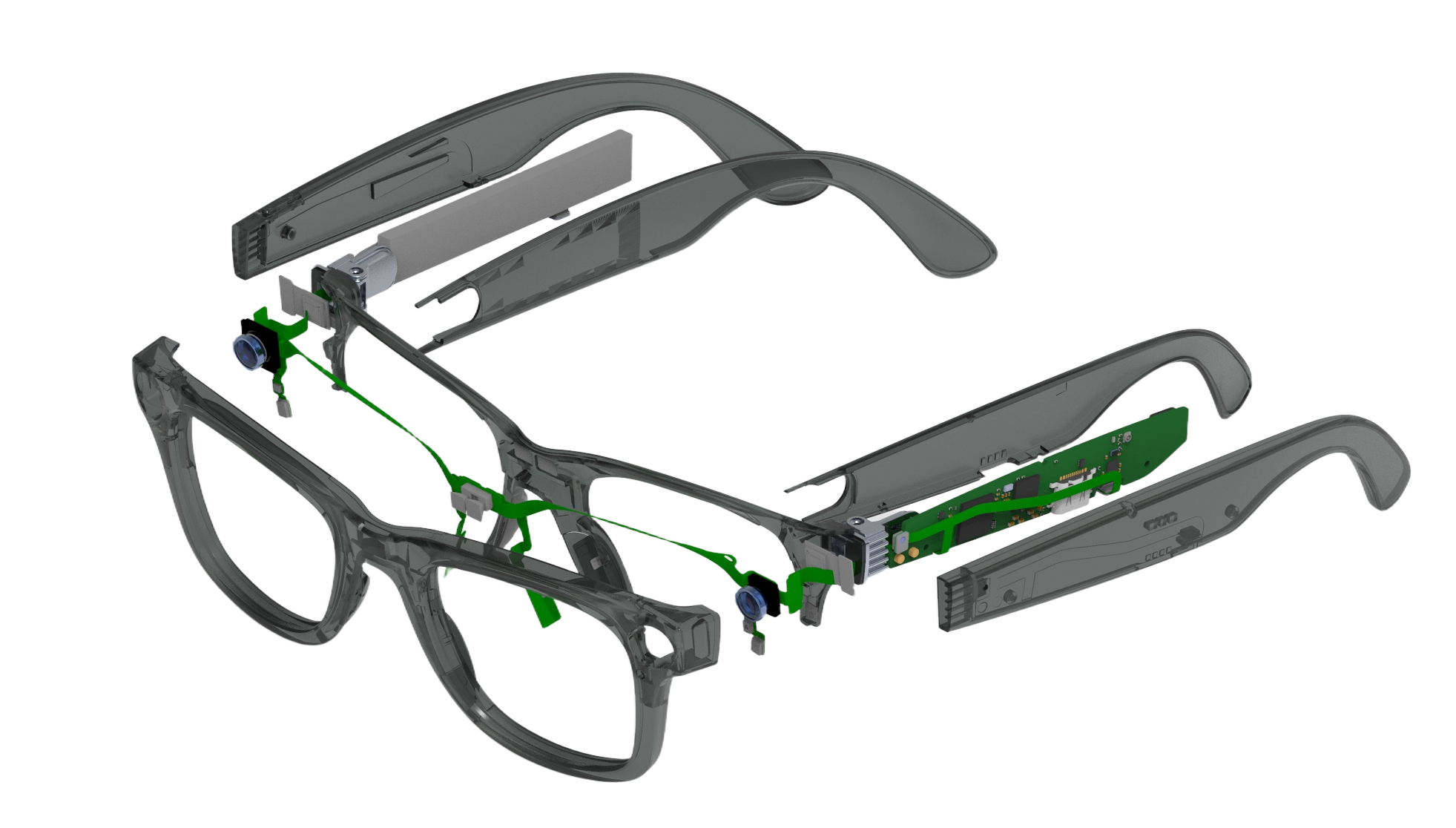}
\caption{Render of the ARGO smart eyewear platform.}
\label{fig:system_render}
\end{figure}

Notably, the embedded hardware Neural Processing Unit (NPU) enables high-efficiency inference locally, significantly reducing the latency typically associated with cloud-dependent systems and preventing the exposure of acquired data to external servers, enhancing the privacy of the user and of the surroundings~\cite{Wang_2025}. To facilitate advanced research across multiple CV domains, the platform integrates a comprehensive suite of multimodal sensors. This sensorized framework is designed to support a wide range of applications, including real-time object detection, gesture recognition, human pose estimation, and Simultaneous Localization and Mapping (SLAM)~\cite{app15147905, SamitaBhandari2025}.

In this article, we provide a comprehensive description of the system, covering the hardware design, firmware implementation, and neural model deployment, as well as its experimental characterization. Particular emphasis is placed on its use in an object detection application based on the YOLOv11 model~\cite{redmon2016you}.

The system is capable of processing image-stream-based neural models, achieving a continuous pipeline of image acquisition, data processing, model inference, and post-processing at a rate of up to 10 FPS, while maintaining an operational autonomy of approximately 113 minutes on a single 200 mAh charge.
While the frame rate could be further enhanced through deep firmware-stack optimizations, the current throughput provides an optimal trade-off between temporal resolution and power consumption.
This ensures reliable egocentric scene understanding without compromising the device's thermal and energy budget.

The remainder of this article is organized as follows: Section~\ref{sec:Related work} provides a comprehensive review of the current state-of-the-art in smart eyewear technologies. Section~\ref{sec:System description} details the hardware architecture and sensor integration. Section~\ref{sec:firmware and software development} focuses on the firmware development and the deployment pipeline for the Machine Learning models. Section~\ref{sec:Results} quantifies the experimental performance and power metrics, while Section~\ref{sec:Future development} outlines potential future enhancements and research directions.

\section{Related work}
\label{sec:Related work}
The contemporary interest in wearable technology has driven both major tech companies and research institutions to develop commercial products and research prototypes aimed at exploring the technological potential and social implications of smart eyewear.

Among commercial solutions, the \textit{Ray-Ban Meta Wayfarer} has successfully captured consumer interest by blending an iconic aesthetic with high-end functionality. These glasses have set a new standard for social interaction, enabling seamless photo and video capture while leveraging a cloud-based infrastructure to run Large Language Models (LLMs) for multimodal user interaction. While the \textit{Ray-Ban Meta} excels in bringing AI to the mass market through a "connected" approach, it serves as a powerful motivator for the research community to push the boundaries of what can be achieved directly on-device, without constant reliance on cloud servers.

To address these research needs, several hardware platforms have emerged to provide a foundation for developing new AI models, software architectures, and egocentric datasets. In this context, Meta's \textit{Project Aria}~\cite{engel2023project} represents the gold standard for multimodal data acquisition. Equipped with a comprehensive sensor suite---including multiple cameras, high-precision IMUs, and microphone arrays---Aria has unlocked new frontiers in egocentric vision research. However, this platform remains a closed, "black-box" ecosystem designed primarily for data collection. For researchers, the lack of control over the underlying hardware prevents deep architectural optimization. It precludes the possibility of performing a tight co-design between the model, the firmware, and the hardware constraints, forcing a reliance on off-board processing or power-hungry mobile SoCs for complex AI tasks.

In parallel, the open-source community has attempted to broaden access to smart eyewear with projects such as \textit{OpenSmartGlass}~\cite{tosg_community}. While these initiatives offer transparency, they often rely on general-purpose microcontrollers like the ESP32-S3~\cite{espressif_esp32s3_datasheet}, which lack dedicated neural acceleration. This results in prohibitive latency when running modern models like YOLO. To compensate for this computational inefficiency, such devices are often forced to use bulky batteries (up to 500 mAh) that compromise ergonomics, while offering minimal sensor integration---typically limited to a single microphone, and one camera.

A more sophisticated approach is seen in \textit{Frame} by Brilliant Labs~\cite{brilliantlabs2024}, which utilizes FPGAs to achieve a sleek form factor. However, Frame is primarily designed as a "companion device" for smartphones, lacking the raw on-device computational throughput required for standalone, high-frequency neural inference. Their recently announced product, the \textit{Halo}~\cite{brilliant_halo_2024}, appears to be a promising upgrade to the Frame platform. Nevertheless, preliminary previews suggest that it still follows a "thin-client" philosophy, offloading acquired data to remote servers for LLM-based question-and-answer interactions rather than performing intensive local vision processing.

An alternative approach emphasizing gesture recognition is exemplified by \textit{Helios}~\cite{bhattacharyya2024helios}, a system integrating an event-based camera into a commercial Ray-Ban Meta frame. In this architecture, data is streamed to an external development board for processing. The primary contribution of this work lies in the efficiency realized through hardware-software co-design. By deploying a hardware-tailored model on an NXP processor, the system achieves an average power consumption of 16 mW, a figure derived from temporal usage patterns typical of smartwatch-style interactions. With a compact 200 mAh battery (providing approximately 0.74 Wh at 3.7 V), this low-power draw enables an estimated operating life of 46 hours. Consequently, these results demonstrate that full-day battery autonomy for "always-on" gesture detection is attainable when hardware, firmware, and AI models are deeply co-optimized.

The research landscape for smart eyewear and egocentric vision systems is decisively shifting toward edge-based processing; however, critical challenges regarding battery autonomy and mechanical integration remain unresolved. While numerous works have attempted to deploy computer vision models on wearable platforms, they often fall into the trap of a disjointed design approach that overlooks the fundamental synergy between hardware, firmware, and AI.

A significant contribution to software optimization is provided by \textit{Boussihmed et al.}~\cite{boussihmed2025tinyml}, who introduced a TinyML model for sidewalk obstacle detection. By scaling a YOLOv5n architecture down to a mere 1.93 MB footprint, the authors achieved a $mAP_{50-95}$ of 22.8\% with a reported inference speed of 96.2 ms. However, this work exemplifies a common gap in non-co-designed systems: validation was conducted on a standard desktop-grade CPU. The absence of a native implementation on ultra-low-power microcontrollers prevents a realistic assessment of energy consumption in a wearable context, leaving the practical feasibility of their 1.93 MB model unquantified in terms of Joules per inference.

The physical integration bottleneck is even more evident in the "standalone" system proposed by \textit{Poy and Darmaraju}~\cite{poy2024standalone}. Their architecture relies on a Raspberry Pi Zero 2 W paired with an external Intel Neural Compute Stick 2 (NCS2). While the use of a YOLOv8-medium model (640×640 input) enables high detection accuracy---achieving a $mAP_{50-95}$ of $63.5\%$ across 270 classes---the dependence on external USB accelerators and a massive 3000 mAh battery results in a bulky, non-ergonomic form factor. Given current lithium-polymer technology, such a battery would occupy approximately 25 $cm^3$---a volume entirely incompatible with the ergonomic constraints of smart eyewear. This footprint, coupled with a lack of reported energy metrics, underscores how sacrificing form-factor for raw computational power remains unavoidable in the absence of a hardware-software co-design centered on integrated neural acceleration.

Following this integrated paradigm, we introduce the ARGO platform. Unlike systems relying on external accelerators or general-purpose processors, ARGO leverages the integrated NPU of the STM32N6~\cite{st_stm32n6_datasheet} to execute an optimized YOLOv11 model at 10 FPS. By embedding this microcontroller into a custom-designed hardware architecture, we overcome the form-factor limitations typical of modular systems built from multiple bulky components. Furthermore, this integration enables a significant reduction in power consumption by optimizing energy delivery to specific subsystems, thereby eliminating the overhead and inefficiencies inherent in general-purpose designs. ARGO provides a holistic solution that balances a rich multimodal perception suite (RGB, Event-based, ToF, IMU, microphones and ambient sensors). By achieving a real-world autonomy of approximately 113 minutes with a battery footprint of only 200 mAh, ARGO bridges the gap between high-fidelity environment sensing and the ergonomic requirements of daily-wear assistive technology.

\section{System description}
\label{sec:System description}
\begin{figure}[t]
    \centering
    \includegraphics[width=\linewidth]{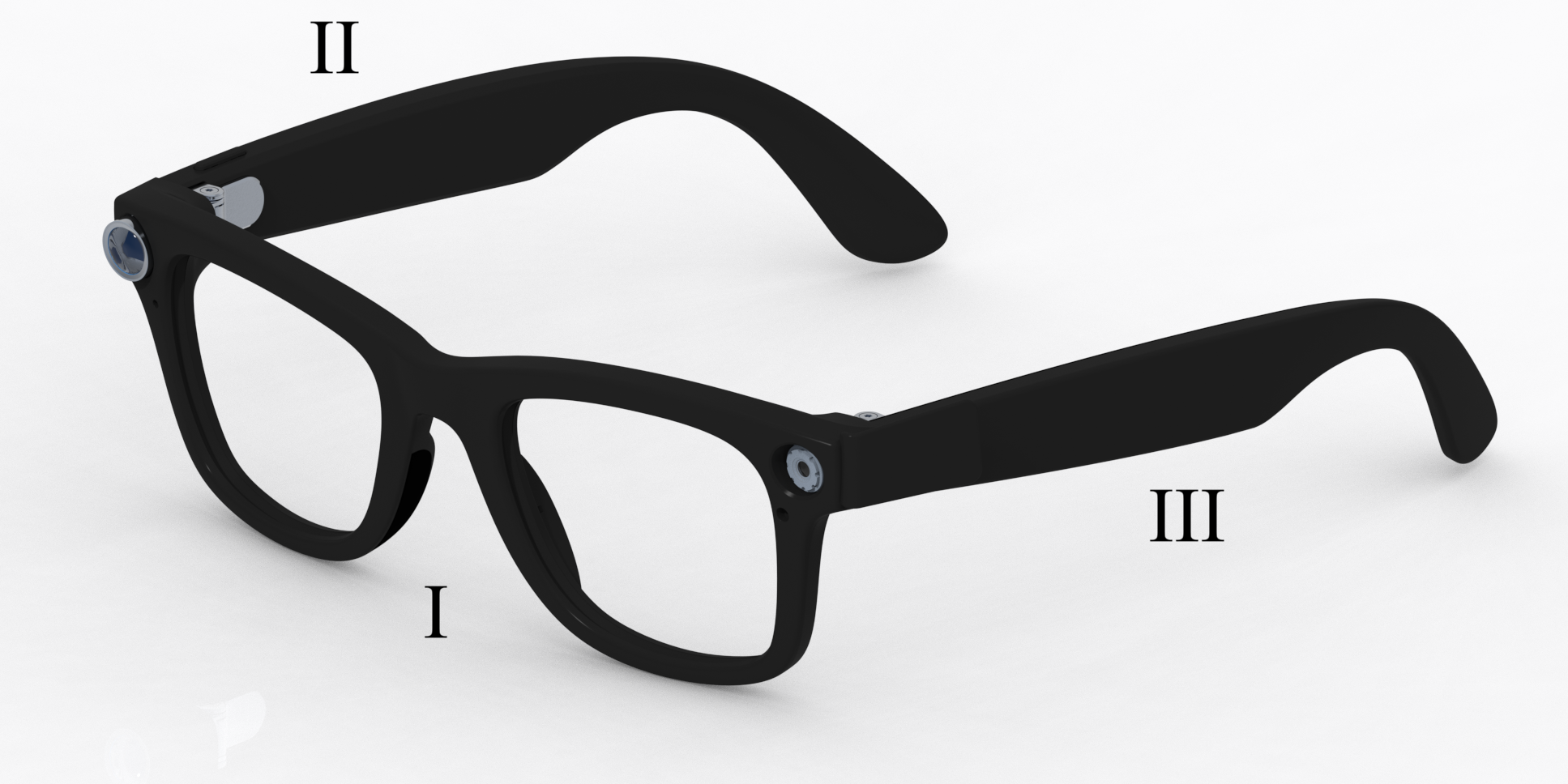}
    \caption{Render of the smart eyewear. Part I comprises the frontal frame, housing a flex PCB, cameras, ToF sensor, microphones, and an ambient light sensor. Part II contains the battery, a pair of microphones, and a push button, while Part III hosts the main board and an additional pair of microphones.}
    \label{fig:esploso_occhiale}
\end{figure}

Given the geometry of the eyewear, the space is divided into three macro-sections: (i) the front frame and the nose pad, which host a sensorized flexible board and two small flexible boards on the nose pad; (ii) the right temple, which houses the battery and two boards, a flexible one hosting two microphones and a rigid interconnection board between the parts; and (iii) the left temple, where the main board resides together with a flexible board hosting two microphones.
Fig.~\ref{fig:esploso_occhiale} shows the high-level schematic of the smart eyewear.
Each section is described in detail below.

\subsection{Frontal and Nose Board}
The frontal board embeds the majority of the sensors. The egocentric location of the frame front is favorable for embedding sensors capable of capturing information from the surrounding environment, such as image sensors, microphones, ToF module, and environmental sensors.
The board is fabricated using flexible-circuit technology and features a six-layer stack-up to allow routing of signals and power traces. The final thickness is approximately 350 µm, enabling integration of the board within the eyewear. Although air-gap technologies would be beneficial to overcome repetitive mechanical stress at the hinges, it was not adopted in this prototype in order to reduce fabrication time and cost.
Communication between this board and the main board integrated in the left temple is achieved through a 60-pin board-to-board connector with a low mating profile (0.6 mm). On the opposite side, a lower-pin-count (26 pins) connector links the right temple, enabling transmission of the battery power line, the push button signal, and the remaining two microphones. Finally, two small connectors (10 pins) positioned at the center of the frontal frame link the frontal board to the two nose boards.

The board was developed in two distinct variants, tailored to specific application requirements through different camera configurations. The first variant features two low-resolution ($<1$ MP) monochrome global-shutter cameras, specifically integrated to support stereo SLAM applications. To ensure temporal alignment, both sensors share an external clock and are triggered synchronously by the processing unit. The second variant hosts a 1.5 MP RGB camera alongside a 0.1 MP event-based sensor. In both configurations, the choice of low-resolution sensors with a high Field Of View (FOV) is strategically dictated by the need to minimize the system's power envelope. High-resolution sensors necessitate elevated internal clock frequencies for data readout, leading to increased power consumption and higher bandwidth demands during transmission and pre-processing. Furthermore, since the visual data is typically downsampled through binning or cropping to match the neural network's input dimensions, the overhead of a high-resolution sensor would provide diminishing returns while incurring a significant energy penalty. This design choice ensures an optimal hardware-software alignment, focusing computational resources on meaningful feature extraction over redundant pixel processing.

With the exception of the event-based sensor, which utilizes a dedicated pinout and connector, all camera modules share a standardized physical interface.
This hardware compatibility enables easy replacement of cameras characterized by different FOVs, pixel architectures, resolutions, and shutter types, making the platform highly versatile.
From a communication standpoint, standard cameras interface with the processing unit via a MIPI CSI-2 bus, with one or two data lanes for data transfer, and an I\textsuperscript{2}C bus for command signaling, whereas the event sensor employs a parallel interface.
Near each camera, a 6-axis Inertial Measurement Unit (IMU) is mounted rigidly with respect to the camera itself, allowing IMU data to be fused with camera data in applications such as SLAM, or to allow for motion artifact correction algorithms.

In both versions, the cameras are placed at the eyewear endpoints, to maximize scene coverage while maintaining a suitable overlap angle between the two cameras. At the eyewear bridge, a ToF sensor is installed, intended for gesture-recognition and depth-mapping applications~\cite{10636485}.

A total of six microphones are distributed across the frontal frame.
Specifically, two vibration microphones are integrated into the nose pad, while the remaining four are mounted on the rims near the rivets. These microphones are arranged in stereo pairs and, with the exception of one pair, share a common reference clock.

In addition to the main sensing array, the frame front hosts ambient light, temperature, and pressure sensors. The parameters they monitor generally vary slowly, aside from abrupt context shifts like moving from a heated indoor space to the outdoors. Thanks to their intrinsic design and minimal data throughput, these components deliver useful contextual insights with virtually zero impact on the system's power and computational budgets~\cite{magno2016infinitime, 10943217}.

\subsection{Temple Boards}

The right temple is the simplest from an electronic standpoint. This section hosts a small rigid interconnection board linking the frontal board previously described, a flexible board with two microphones, and the battery. A push button is also placed on this rigid board and can be actuated through a metal button located on the upper part of the eyewear temple and operated by the user.
Most of the available space is occupied by the 200 mAh lithium battery. The two-layer flexible board, embedded in the plastic and routed behind the battery, carries the interconnections of the two microphones configured as a stereo pair.
The left temple hosts the main board, which contains the entire power, processing, and communication sections, the latter relying on a Bluetooth Low Energy (BLE) microcontroller enabling wireless data streaming to external host devices. The board is further complemented by an analogous flexible board accommodating two additional microphones in the same stereo configuration, embedded within the plastic to maximize the available space.

Each temple thus contributes two microphones, for a total of four, which, combined with the six located on the front frame, yields ten microphones overall. This number intentionally exceeds the six supported by the processing unit: given the lack of prior knowledge regarding optimal microphone placement, the additional units enable evaluation of the most effective positioning strategy for a potential second prototype.

The main board is fabricated on an 8-layer rigid stack-up, with a final thickness of 1 mm. Given the limited available space, the board employs laser-drilled blind and buried microvias to optimize component placement and the routing of the most critical signal-integrity nets.

The board includes a USB Type-C connector that enables system power supply, battery charging, and data transfer to and from a host computer. In the absence of a USB-C cable, the system is powered by the battery located in the right temple.

System power-up is controlled by the state of a slide switch accessible from the inside of the left temple, which acts as the enable signal for the mounted Power Management Integrated Circuit (PMIC).
The selected PMIC integrates multiple functionalities, including battery management compliant with the Japan Electronics and Information Technology Industries Association (JEITA) standard, state-of-charge monitoring through an integrated fuel gauge, and voltage regulation via three integrated buck-boost converters and two Low Dropout (LDO) regulators.

\begin{figure*}[t]
\centering
\includegraphics[width=0.85\linewidth]{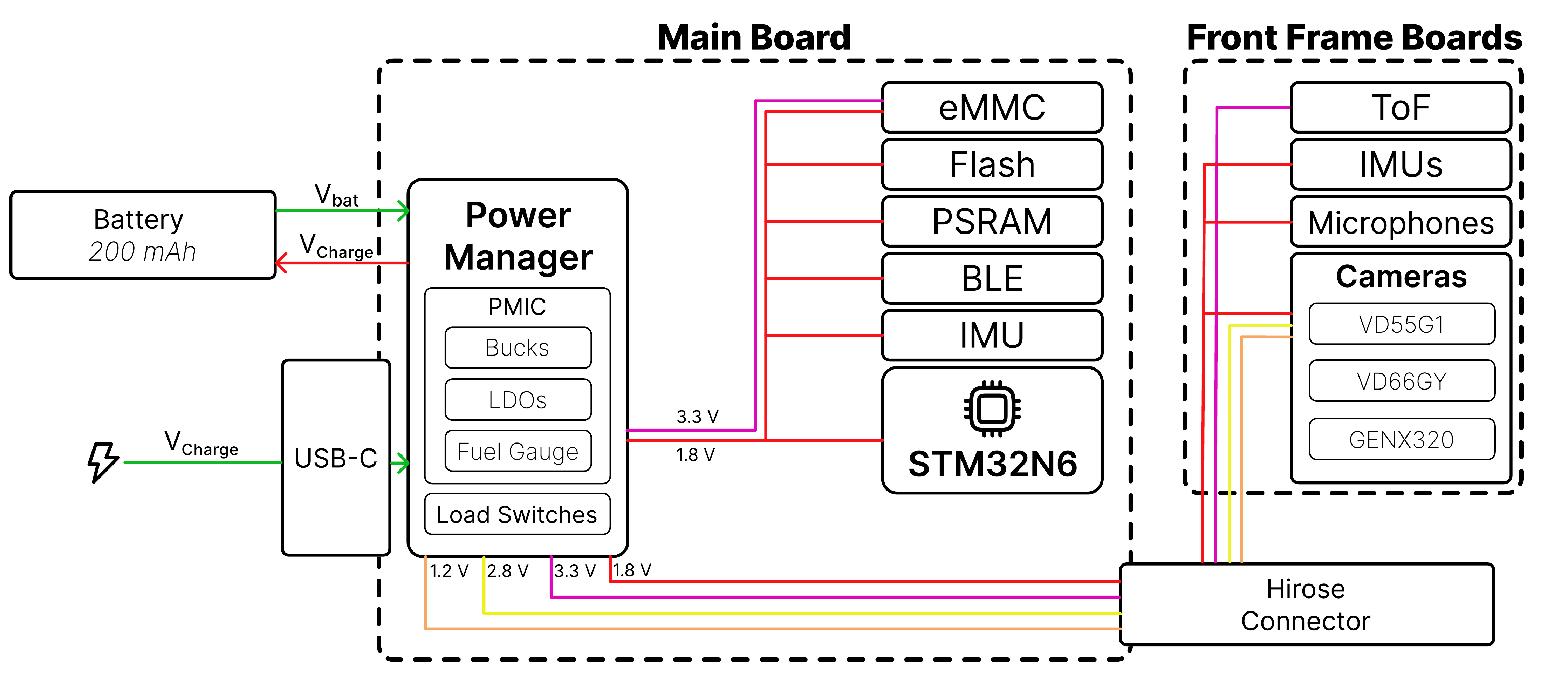}
\caption{Power scheme of the whole system. The core of the power supply is provided by the PMIC.}
\label{fig:schema_potenza}
\end{figure*}

Finally, the PMIC adopts a Single-Inductor Multiple-Output (SIMO) topology, allowing the use of a single inductor, thereby saving space and reducing the size of the power-stage circuitry.
The power schematic is shown in Fig.~\ref{fig:schema_potenza}.

Starting from the battery voltage---$4.2~V$ when fully charged and down to $2.9~V$ when discharged---or from the fixed 5~V supplied by USB, the PMIC's buck converters generate the $1.8~V$, $3.3V_{main}$, and $3.3V_{aux}$ rails.
From these, the two LDOs provide the $2.7~V$ rail (derived from $3.3V_{main}$) and a $1.8V_{LDO}$ rail derived from $3.3V_{aux}$. These two LDO-filtered voltages are used to supply the system's most critical components, namely the sensor suite, the analog section of the processing unit, and the BLE microcontroller.
Finally, to generate the correct supply voltages for the cameras used in the two PCB versions mounted on the eyewear front module, a small WLCSP-package buck converter is used to derive $1.15~V$ from the $1.8V_{LDO}$ rail, and two power switches are employed to guarantee, together with the buck enable signal, the correct power-up sequence of the front-mounted cameras.
To ensure the correct power-up sequence of the various voltage domains inside the processing unit, two small power switches are driven directly by enable signals originating from the processing unit itself. Finally, to optimize system efficiency, the input voltage of the LDOs is adjusted to remain just above the threshold defined by the sum of the output voltage, the dropout voltage ($V_{dropout}$), and a safety margin. Minimizing the voltage drop across the LDO drastically reduces thermal power dissipation, thereby enhancing the overall power conversion efficiency of the system.

The main processing unit is the STM32N6 MCU (STMicroelectronics), selected to satisfy the strict form-factor constraints of the eyewear while delivering the required throughput for both general-purpose tasks and neural inference.
To support this dual workload, the device integrates 4~MB of internal SRAM, allocated between the CPU---capable of reaching 900~MHz---and the NPU, which can operate up to 1~GHz, achieving up to 0.6~TOPS.
To fully exploit the computing capabilities of the MCU, it is complemented by an external 64~MB PSRAM and a 128~MB external FLASH memory, used for storing the application as well as the weights and activations of neural models when their size exceeds the available on-chip memory.
To fit within the eyewear frame, the selected MCU package measures $6~mm \times 6~mm$ and provides a single OctoSPI communication interface, shared by both external memories. This configuration requires additional care during neural model deployment, since only one of the two memories can be configured in memory-mapped mode.
In this mode, the external memory is mapped directly into the microcontroller's internal address space, allowing the CPU and NPU to access data transparently without executing low-level SPI commands.
Specifically, if a model necessitates both external volatile and non-volatile memories simultaneously, the system incurs a significant software-management overhead to manually route data to and from the non-mapped component.

For low-speed peripheral control, the system uses four distinct I\textsuperscript{2}C buses.
A primary bus connects the non-imaging sensors, while two dedicated buses independently serve the cameras.
The fourth I\textsuperscript{2}C interface is shared by the N6 MCU, the BLE device, and the PMIC to facilitate dynamic power adjustments.

For external communication, the system uses the USB OTG 2.0 High-Speed interface to transfer data to a host PC, or it sends data via UART to the auxiliary BLE MCU that implements the BLE stack.
This auxiliary MCU manages the bidirectional communication between a smartphone BLE application and the primary MCU. It facilitates the execution of demonstrations and data acquisition sessions while routing feedback data for real-time visualization. By offloading the BLE stack management to this dedicated unit, the main MCU can allocate its full computational resources to the core sensing application.
Table~\ref{tab:system_buses} summarizes the communication buses used in the system.

\begin{table}[!t]
\caption{System communication interfaces and bus protocols.}
\label{tab:system_buses}
\centering
\renewcommand{\arraystretch}{1.1}
\setlength{\tabcolsep}{6pt}
\begin{tabular}{ll}
\toprule
\textbf{Component} & \textbf{Bus Interface} \\
\midrule
Cameras                  & MIPI CSI-2, parallel, I\textsuperscript{2}C \\
Microphones              & PDM \\
BLE module               & UART \\
Flash and PSRAM          & OctoSPI (shared bus) \\
USB port                 & USB 2.0 HS \\
IMUs, PMIC, ToF, sensors & I\textsuperscript{2}C \\
\bottomrule
\end{tabular}
\end{table}

Finally, we used a precision scale to measure the weight of each individual component. We found that the electronics housed in the left temple have the same weight as the combined battery and circuitry in the right temple. Furthermore, the total weight of the electronic subsystem---including the battery---accounts for less than 20\% of the device's overall mass, which is approximately 41 grams (excluding lenses).

\section{Firmware and Software Development}
\label{sec:firmware and software development}
To validate the proposed architecture and demonstrate the necessity of a holistic co-design approach for edge AI, we developed a case study focused on detecting hazardous urban objects.
Rather than treating hardware, firmware, and machine learning as isolated domains, this application illustrates an end-to-end pipeline tightly optimized for the platform's specific constraints.
Specifically, the ARGO platform was enhanced with a dedicated deployment pipeline for a customized YOLOv11 detector~\cite{jocher2023yolo}, targeting real-time object recognition via the frontal RGB camera board.

Ensuring the seamless operation of the system and the execution of the neural model required the development of dedicated application firmware. The code was implemented in a bare-metal environment to simplify the coding stage and avoid excessive Real-Time Operating System (RTOS) overhead. The system's firmware architecture is divided into two primary macro-functions: a setup function for system initialization and an application execution function.

During the setup phase, the MCU configures its internal peripherals, sensors, and external components, including the RGB camera and external Flash memory in memory mapped mode. Additionally, it manages the voltage rails to ensure the correct power-up sequence for the various board sections. Following initialization, the system enters a standby, yet not low-power, mode, awaiting the application's start signal, which can be triggered either via a BLE command or a physical pushbutton located on the right temple of the eyewear.

\begin{figure}[t]
    \centering
    \includegraphics[width=0.7\linewidth]{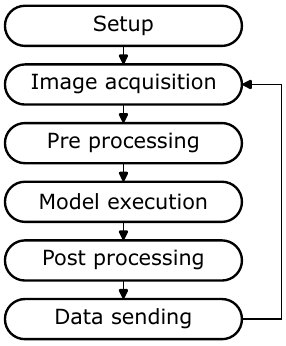}
    \caption{High-level representation of the system's Finite State Machine (FSM). Following an initial setup phase, the system awaits a start command from the application; subsequently, it cycles through all operational states in a continuous loop.}
    \label{fig:schema_applicazione}
\end{figure}

Once initialized, the application executes the state machine illustrated in Fig.~\ref{fig:schema_applicazione}.

The application workflow begins with the N6 MCU acquiring an image from the camera and applying pre-processing algorithms via its Digital Camera Memory Interface and Pixel Processor (DCMIPP) peripheral. Subsequently, the image is rescaled and formatted as input for the neural network. After performing inference, the network's output is processed and transmitted via UART to the BLE MCU, which encapsulates the data packet and forwards it to the smartphone. Alternatively, for debugging purposes, the data stream can be routed to a computer via USB. Finally, the smartphone application notifies the user of specific objects detected in the surrounding environment. Note that the development of the mobile application is beyond the scope of this paper and will not be described.

The key components of this integrated execution flow, encompassing both hardware acceleration and firmware management, are detailed below.

\subsection{Camera Acquisition}
During the system power-up phase, the camera is initialized and configured in snapshot mode. Given the low operating frame rate of approximately 10 Hz, the camera is transitioned into a standby, yet not low power, state between consecutive acquisitions to further reduce the energy footprint~\cite{likamwa2013energy}. The integration phase is externally triggered by the N6 MCU. To ensure robust performance under varying lighting conditions, the camera utilizes an automatic exposure control mechanism. Once the frame acquisition is complete, the data is transmitted to the N6 MCU via the MIPI CSI-2 protocol; subsequently, the MCU drives the camera back into standby mode to conclude the cycle.

\subsection{Microcontroller Pre-processing}
Upon receiving the frame from the camera, the N6 MCU performs pre-processing via its DCMIPP peripheral before forwarding the data to the NPU.
To preserve the original FOV, the initial $1120 \times 1120$ frame is rescaled to match the specific input dimensions required by the neural network.
Theoretically, resizing could be delegated to the camera's internal Image Signal Processor (ISP) via analog pixel binning to reduce I/O traffic and power consumption. However, empirical characterization showed only marginal energy gains, leading us to perform this operation within the MCU to maintain full control over the processing pipeline. In this stage, the N6's ISP applies hardware-accelerated debayering, color correction, and white balance. These adjustments normalize the input data against lighting variations, ensuring consistency with the training dataset and enhancing the robustness of the neural inference.
Moreover, at the DCMIPP output, the frame retains the camera's original channel-last (interleaved) format, where pixels are arranged as $R_1, G_1, B_1, R_2, G_2, B_2, \dots$.
To satisfy the planar (channel-first) input requirements of the neural network, a transformation is applied to reorder the array into an $R_1, R_2, \dots, G_1, G_2, \dots, B_1, B_2, \dots$ structure. Following this pre-processing and reformatting stage, the image is ready for inference, as detailed in the next section.

\subsection{Machine Learning Deployment and Optimization}
\label{sec:modello}

To fully exploit the N6 neural acceleration capabilities while maintaining a low energy footprint, the YOLOv11 model must be meticulously optimized for the underlying hardware.
First, inference must comply with the platform memory organization, including internal SRAM and the NPU local buffers.
Second, the computational graph must be shaped to avoid unsupported operators, which may prevent successful compilation, and to minimize execution outside the NPU, so that inference remains predominantly accelerator-driven.
To meet these requirements, the pipeline is organized into three stages: (i) training multiple YOLOv11 variants on the Walking On The Road (WOTR) dataset, (ii) applying architectural graph modifications to enforce NPU-compatible tensor ranks, and (iii) performing post-training quantization together with output-graph optimizations.

\subsubsection{Dataset and Training}
To train the object detection models on visually representative walking scenarios, we adopted the WOTR dataset~\cite{XIA2023102486}.
Although WOTR is not captured directly from a head-mounted device, its visual content closely reflects what a pedestrian typically observes while walking in urban environments.
This makes it a suitable representative dataset for the ARGO operating context, supporting the learning of obstacles and infrastructure elements that are critical for safe navigation.

The WOTR dataset includes 20 object categories, grouped into three functional classes:
\begin{itemize}
    \item \textbf{Infrastructure and Signage}: \textit{tree, red light, green light, crosswalk, sign, pole}, and \textit{fire hydrant}.
    \item \textbf{Urban Mobility Agents}: \textit{person, bicycle, bus, truck, car, motorcycle, tricycle}, and \textit{dog}.
    \item \textbf{Hazards and Navigational Aids}: \textit{blind road, reflective cone, ashcan, warning column}, and \textit{roadblock}.
\end{itemize}

\begin{figure*}[t]
    \centering
    \begin{minipage}{0.31\textwidth}
        \centering
        \includegraphics[width=\textwidth]{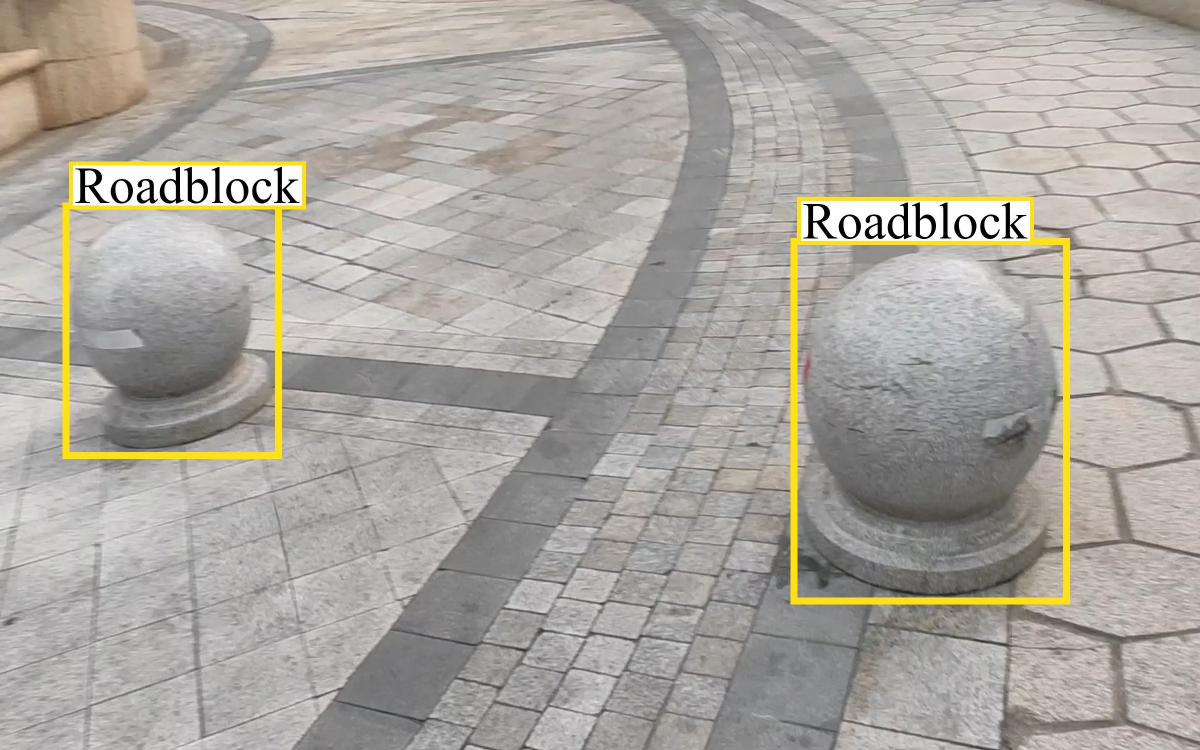}
        \small \texttt{A.}
    \end{minipage}
    \hfill
    \begin{minipage}{0.31\textwidth}
        \centering
        \includegraphics[width=\textwidth]{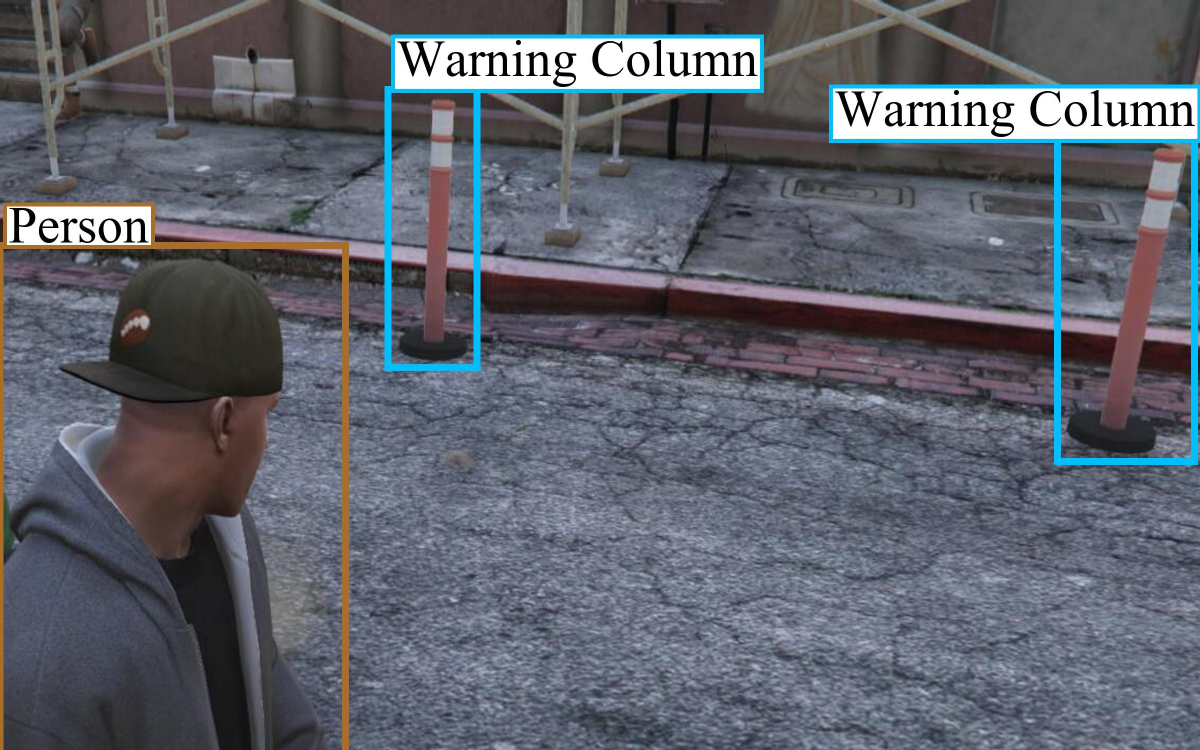}
        \small \texttt{B.}
    \end{minipage}
    \hfill
    \begin{minipage}{0.31\textwidth}
        \centering
        \includegraphics[width=\textwidth]{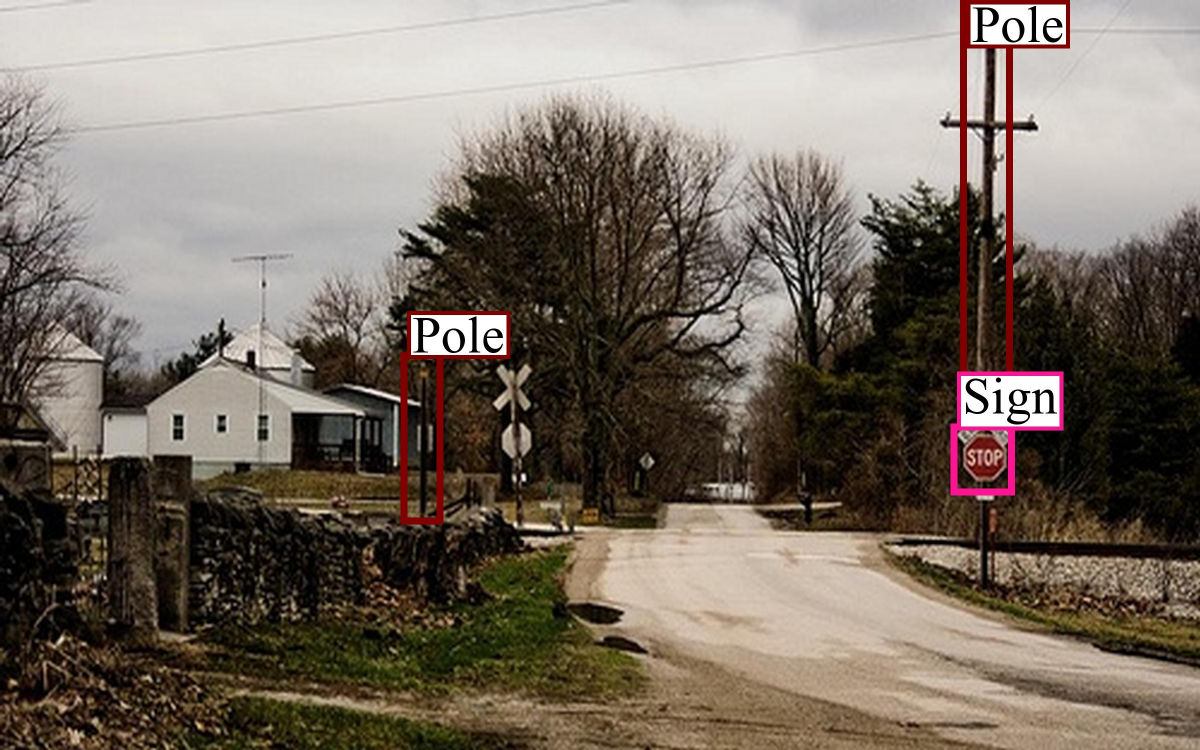}
        \small \texttt{C.}
    \end{minipage}

    \vspace{0.4cm}

    \begin{minipage}{0.31\textwidth}
        \centering
        \includegraphics[width=\textwidth]{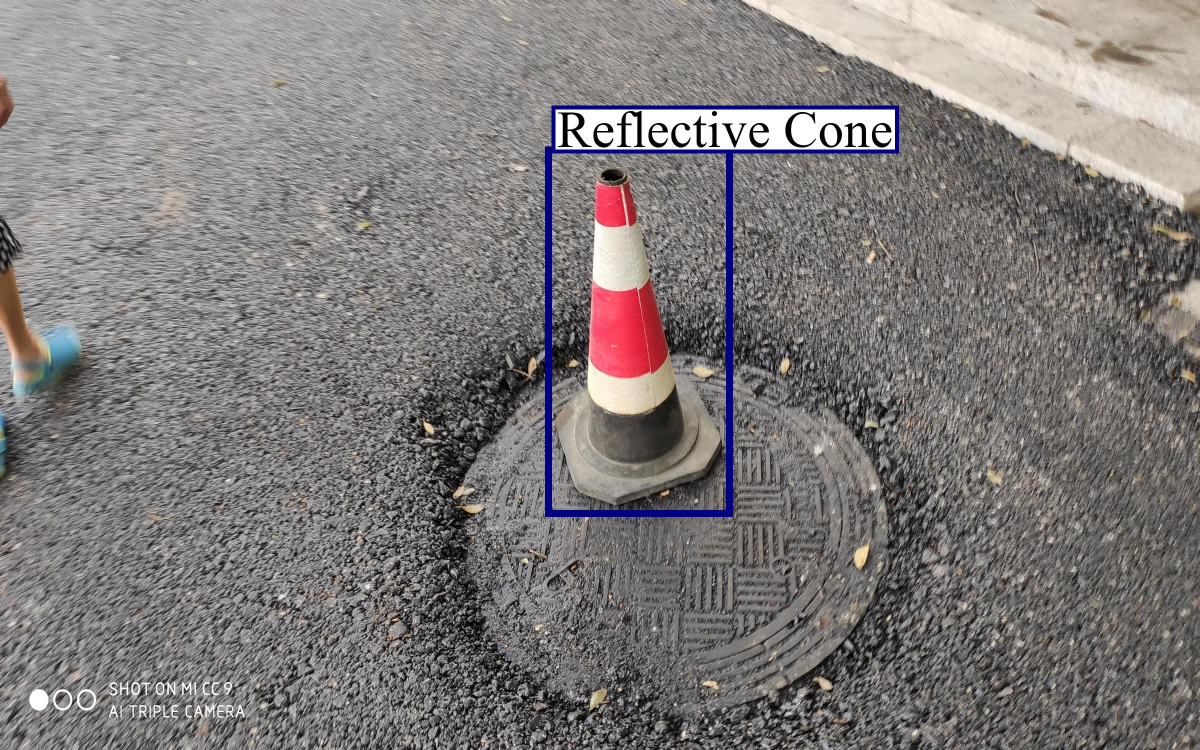}
        \small \texttt{D.}
    \end{minipage}
    \hspace{1.5cm}
    \begin{minipage}{0.31\textwidth}
        \centering
        \includegraphics[width=\textwidth]{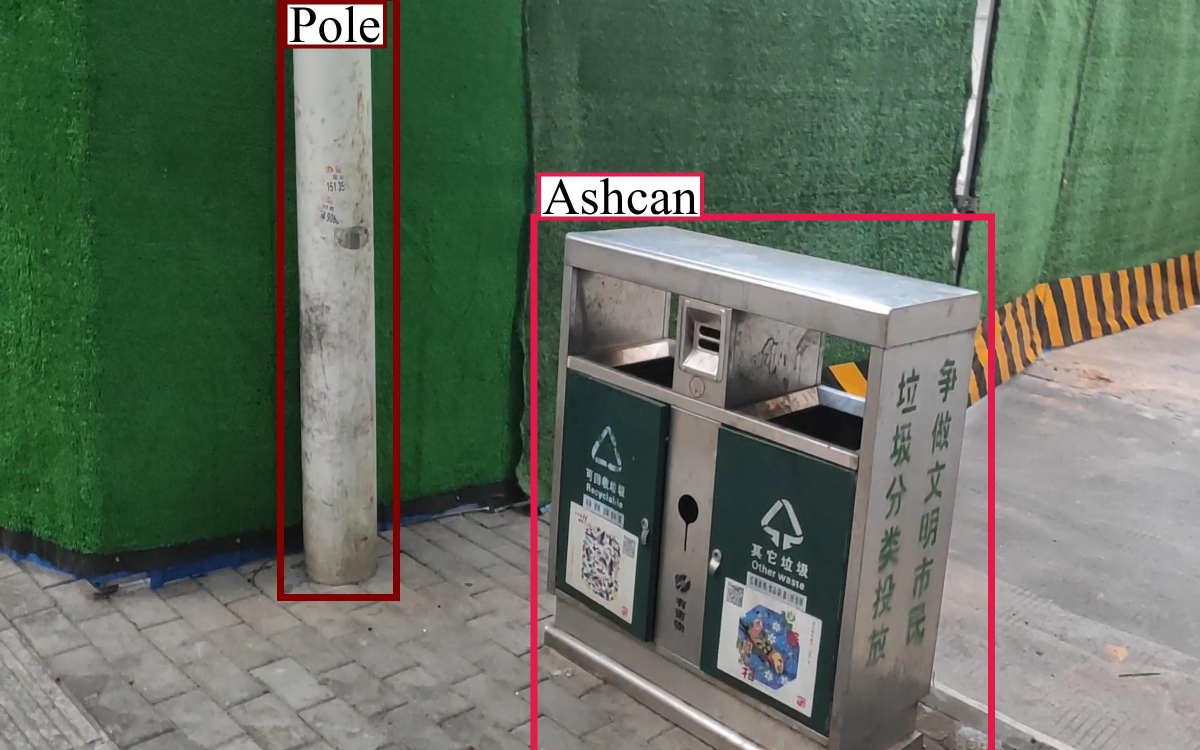}
        \small \texttt{E.}
    \end{minipage}

    \caption{Representative samples from the WOTR dataset. Each panel illustrates typical urban scenes containing pedestrians, vehicles, and infrastructure elements.}
    \label{fig:wotr_samples}
\end{figure*}

Fig.~\ref{fig:wotr_samples} presents representative samples from WOTR, highlighting typical urban scenes and the variety of annotated objects encountered in pedestrian-level views.

Training was performed with the Ultralytics YOLO framework (v8.3.223) using a fine-tuning strategy for the object detection task.
The dataset comprises 13928 images, split into 9056 training images (65\%), 2338 validation images (17\%), and 2534 test images (18\%).

To identify the optimal balance between predictive performance and hardware resource utilization, we evaluated three architecture variants: YOLOv11-\textit{n} (nano), YOLOv11-\textit{s} (small), and YOLOv11-\textit{m} (medium), each tested at four input resolutions (128, 192, 256, and 320 pixels).
This grid was designed to probe two distinct but coupled constraints: (i) the non-volatile footprint of model parameters in FLASH, and (ii) the volatile footprint of intermediate activations in SRAM/NPU local buffers.

\paragraph{Non-Volatile Memory}
The model YOLO variant ($n$, $s$, $m$) determines the network structural complexity (depth/width), hence the number of layers and trainable parameters $P$.
Since weights are stored persistently, the FLASH occupation (\(M_{w}\)) is essentially fixed once the variant is selected, and scales approximately with the parameter count:
\begin{equation}
    M_{w} \propto P
    \label{eq:weights_flash}
\end{equation}
Larger variants were excluded to ensure the final binary remains within the ARGO platform non-volatile storage budget.

\paragraph{Volatile Memory}
Unlike weights, activation tensors are a dynamic inference-time load.
For any fixed model variant, the input spatial resolution $S$ dominates the feature-map sizes throughout the network, and therefore the peak volatile memory requirement.
In first approximation, the activation peak scales quadratically with $S$:
\begin{equation}
    M_{act} \propto \kappa \cdot S^2,
    \qquad
    \kappa = \sum_{l=1}^{L} \frac{C_l}{\sigma_l^2}
    \label{eq:memory_scaling}
\end{equation}
where $C_l$ and $\sigma_l$ denote the number of channels and the cumulative stride at layer $l$, respectively.

The same resolution sweep is also relevant from a computational standpoint, since the dominant convolutional workloads scale with the spatial extent of intermediate feature maps, the number of MACC (Multiply-Accumulate) operations increases approximately as $\mathrm{MACC} \propto S^2$ for a fixed architecture, and therefore impacts inference latency under fixed clock and parallelism.
Accordingly, evaluating multiple input sizes for each variant allows us to quantify the accuracy gains achievable through higher spatial resolution against the corresponding increases in activation memory, compute, and latency.

All models were initialized with pre-trained weights from the COCO dataset~\cite{lin2014microsoft} to improve convergence and stability, leveraging the similarities between general urban scenes and the WOTR domain.

The training procedure was standardized across all runs with a budget of 100 epochs, a batch size of 16, and the Stochastic Gradient Descent (SGD) optimizer. A cosine learning rate schedule was employed to refine the weights, and a fixed seed was set to ensure reproducibility.
To enhance the generalization of the model under variable lighting and urban clutter, a comprehensive data augmentation strategy was implemented.
This included color jittering (HSV), geometric transformations (translation, scaling, and horizontal flipping), and compositional techniques such as mosaic (0.8 probability) and mixup (0.1 probability).
Notably, the mosaic augmentation was disabled during the final 10 epochs to stabilize the bounding box regression before completing the training.

\subsubsection{Structural Adaptation for NPU Acceleration}
The baseline YOLOv11 model includes a Cross-Stage Partial Spatial Attention (C2PSA) block whose Multi-Head Attention (MHA) relies on matrix multiplications defined over 4D tensors.
On the target platform, such 4D matrix multiplication operators are unsupported and violate hybrid execution constraints, causing compilation failures rather than a CPU fallback \cite{ST_NeuralART_NPU}.
Therefore, the attention block was restructured to ensure that the attention-related matrix multiplications can be expressed using supported 3D tensor kernels, enabling successful model deployment and predominantly on-NPU execution.

Let the query, key and value tensors be $Q, K \in \mathbb{R}^{B \times H \times D_k \times N}$ and $V \in \mathbb{R}^{B \times H \times D_v \times N}$, where $B$ is the batch size, $H$ is the number of heads, $N$ is the number of spatial tokens, $D_v$ is the per-head value width, and $D_k{=}0.5D_v$ is the per-head query/key width.
We replace the original 4D MHA with a Head-wise Parallel Attention (HPA) formulation, implemented as $H$ independent attention heads:

\begin{equation}
Q_i, K_i \in \mathbb{R}^{B \times D_k \times N},\quad
V_i \in \mathbb{R}^{B \times D_v \times N}, \quad i=1,\dots,H
\end{equation}

In HPA, each head is computed through a sequence of 3D matrix multiplications only, thereby preventing the runtime failure. For each head, the attention score tensor is computed as:
\begin{equation}
    S_i = Q_i^T \cdot K_i \quad \in \mathbb{R}^{B \times N \times N}
    \label{eq:attn_score}
\end{equation}
followed by scaling and softmax normalization to obtain attention weights $A_i$. The head output is then computed as:

\begin{equation}
    O_i = V_i \cdot A_i^T \quad \in \mathbb{R}^{B \times D_v \times N}
    \label{eq:attn_output}
\end{equation}

Finally, all head outputs $\{O_i\}_{i=1}^H$ are concatenated along channels. As in the baseline block, the concatenated map is added to a depthwise $3\times3$ positional encoding of the value tensor, followed by the unchanged $1\times1$ projection.
This preserves the functional role of the original attention module while ensuring that the computational graph is composed of NPU-friendly operators (3D matmul, elementwise scaling, softmax, concatenation, and linear projection), thus maximizing on-accelerator execution.
Because only the evaluation layout changes, the reformulation introduces no additional modeling error beyond floating-point evaluation order.

\subsubsection{Quantization Strategy}
To reduce memory footprint and improve execution efficiency, the model was quantized using Post-Training Static Quantization (PTQ) with a QuantizeLinear/DequantizeLinear (QDQ) graph format.
Quantization was applied with distinct schemes for weights and activations, chosen to preserve accuracy under urban-scene statistics.

\paragraph{Weights}
All convolution and linear weights were quantized using symmetric per-channel int8 quantization. For each output channel, the scale factor $s_w$ was computed from the maximum absolute weight value, and weights were mapped as:
\begin{equation}
    w_{q} = \text{clamp}\left(\text{round}\left(\frac{w}{s_w}\right), -128, 127\right)
    \label{eq:quant_weights}
\end{equation}
Per-channel quantization was selected to mitigate inter-filter dynamic-range variability, a common source of accuracy degradation under uniform per-tensor mapping.

\paragraph{Activations}
Activations were quantized with an asymmetric scheme. Calibration ranges were estimated with a Percentile method, in order to reduce sensitivity to outliers frequently observed in urban sensing data (e.g., specular highlights, abrupt exposure changes).

Operationally, activation histograms were collected over a representative calibration set from WOTR train set, and the quantization range was set by clipping extreme tails according to the chosen percentile threshold (i.e., 99.999\%).

\subsubsection{Terminal Node Optimization and Deployment}
To streamline the interface between the NPU-executed graph and CPU-side post-processing, we simplified the exported QDQ graph at the output boundary.
Specifically, we removed the final Quantize/Dequantize pair after the sigmoid applied to the class-score branch.
Since sigmoid outputs are naturally bounded in $[0,1]$, quantizing this terminal activation introduced avoidable precision loss on small score differences, which in turn degraded the ranking and thresholding steps used to form the final detections.
By exposing the sigmoid output directly in float32, we preserve score fidelity at the model boundary while keeping the remaining graph quantized.
The final model was compiled using ST Edge AI (v4.0.0), which generates the C code and schedules the operators across NPU, hybrid, and CPU execution domains.
To validate the deployed implementation prior to the on-device characterization, numerical agreement between the host-side reference execution and the STM32N6 execution was evaluated over the full WOTR test set ($N=2534$).

\subsection{Model Post-processing}
\label{sec:postprocess}

The outputs produced by a deployed YOLOv11 detector are not directly the final detections used by the application.
Instead, the network head emits a dense set of raw predictions, which must be decoded on-device into a bounded list of final bounding boxes suitable for downstream communication and feedback.

The head output can be represented as a tensor of shape $B \times (4 + C) \times N$, where $B$ is the batch size (inference is performed with $B=1$), $C$ is the number of target classes ($C=20$ in this work), and $N$ is the number of raw predictions produced for the input frame.
For each prediction $i \in \{1,\dots,N\}$, the $(4+C)$ channels concatenate the 4 bounding-box parameters with the $C$ per-class scores.

For a square input of size $S \times S$, $N$ depends on the output resolution levels of the detection head.
The total number of prediction locations is
\begin{equation}
N(S) = \left(\frac{S}{8}\right)^2 + \left(\frac{S}{16}\right)^2 + \left(\frac{S}{32}\right)^2
\label{eq:num_predictions}
\end{equation}
with $S$ constrained to multiples of 32 in our experiments.
Accordingly, $N(128)=336$, $N(192)=756$, $N(256)=1344$, and $N(320)=2100$ raw predictions are produced per frame prior to any post-processing.

Finally, for each prediction $i$, class and confidence are obtained by argmax over the 20 class scores:
\[
\text{cls}_i=\arg\max_c s_{i,c}
\qquad
\text{conf}_i=\max_c s_{i,c}
\]
and candidates with $\text{conf}_i < \tau_{\text{conf}}$ are discarded.

The remaining candidates may still exceed the bandwidth and latency budget of the embedded-to-phone BLE link; therefore, ARGO introduces a priority-based selection stage that limits the number of transmitted detections while preserving those most relevant for user safety.
Specifically, each candidate is assigned a priority score:
\begin{equation}
    \text{priority score}_i = \text{conf}_i + \alpha \cdot P[\text{cls}_i]
    \label{eq:priority_score}
\end{equation}
where $\text{conf}_i \in [0,1]$ is the detector confidence, $P[\cdot]$ is an application-defined lookup table over the 20 classes, and $\alpha$ controls the contribution of the class-dependent term.
To define $P[\cdot]$ (i.e., class priors) we used integer values ($1$--$5$), whereas the resulting priority score in Eq.~\eqref{eq:priority_score} is continuous and jointly determined by $\text{conf}_i$ and $\alpha$;
In our application, we set $\alpha=0.1$; with $\text{conf}\in[0,1]$, the additive term $\alpha\cdot P[\text{cls}]$ therefore lies in $[0.1,0.5]$.
We instantiate the lookup table as follows: $P[\text{class}] = 5$ for \{bus, car, person, truck\}, $P[\text{class}] = 4$ for \{bicycle, crosswalk, dog, motorcycle, tricycle\}, $P[\text{class}] = 3$ for \{green light, red light, reflective cone, roadblock, warning column\}, $P[\text{class}] = 2$ for \{blind road, fire hydrant, pole, sign\}, and $P[\text{class}] = 1$ for \{ashcan, tree\}.
If the number of valid candidates exceeds the budget $N_{\max}$ (i.e., 20), candidates are sorted in descending priority-score order and truncated to the top-$N_{\max}$ detections.
This prioritization reflects an application-driven risk trade-off: it is preferable to preserve detections from high-risk classes (e.g., vehicles) even at the cost of discarding lower-risk objects, since missing a potentially hazardous object (false negative) is more critical than omitting a benign one (e.g., ashcan).

Finally, redundancy is reduced using Non-Maximum Suppression (NMS).
When truncation is triggered, candidates are first sorted in descending \textit{priority score} order and then processed by a class-aware NMS, so that higher-priority candidates are evaluated first and can suppress lower-ranked overlaps within the same predicted class.
Overlap between two predicted bounding boxes $b_i$ and $b_j$ is measured using Intersection-over-Union (IoU), defined as the ratio between their intersection area and union area:
\begin{equation}
  \text{IoU}(b_i,b_j)=\frac{|b_i \cap b_j|}{|b_i|+|b_j|-|b_i \cap b_j| + \epsilon},
\quad \epsilon=10^{-6}
\end{equation}
where $|b_i|$ and $|b_j|$ denote the areas of $b_i$ and $b_j$, $|b_i \cap b_j|$ denotes the area of their overlap, and $\epsilon$ is a small constant used for numerical stability to avoid division by zero (e.g., for degenerate boxes).
Finally, candidates with $\text{IoU}(b_i,b_j)>\tau_{\text{IoU}}$ are suppressed.

\paragraph{Accuracy evaluation}
All reported detection metrics are computed on the final set of post-processed detections produced by the pipeline described above, i.e., after confidence-based candidate filtering, optional truncation to a fixed detection budget, and class-aware non-maximum suppression.
Evaluation uses a low confidence threshold, an IoU threshold of $0.6$ for suppression, and a maximum of 300 detections per image.
This evaluation setting does not enforce the stricter detection budget used in the deployed application, so as to avoid limiting detector performance during accuracy assessment.
Detection accuracy is reported in terms of \(mAP_{50\text{--}95}\) on the WOTR test split (2{,}534 images), where the IoU criteria used for metric computation follow the standard COCO-style definition and are therefore distinct from the IoU threshold adopted within NMS.
For the quantization analysis, each YOLOv11 variant (n/s/m) and input size $\{128,192,256,320\}$ is evaluated under identical settings in FP32 and after INT8 PTQ, so that any observed difference reflects the effect of post-training quantization under a fixed post-processing and evaluation protocol.

\subsection{Data Communication}
Once the post-processing phase is complete, the data is transmitted to a secondary MCU responsible for BLE communication via a UART serial interface. To optimize power consumption, this MCU remains in a standby and low-power state until UART data is received, while simultaneously maintaining the stability of the BLE connection. Once processed, the resulting data is encapsulated into packets and transmitted via BLE to a smartphone or any compatible terminal for parsing.
Notably, the mobile device receives only high-level detection metadata---specifically, the final post-processed detections encoded as class labels, confidence scores, and bounding-box coordinates---rather than raw image frames.
This design choice is necessitated by both BLE bandwidth constraints and a privacy-by-design approach, ensuring that sensitive visual data never leaves the local edge processing unit.

Beyond BLE transmission, the system supports High-Speed USB communication at 480 Mbps, providing sufficient bandwidth for the concurrent transfer of raw image frames and their corresponding inference results. Upon reception by a host PC, a Python-based utility performs real-time visualization by overlaying the processing metadata onto the video stream. This high-bandwidth link is intended exclusively for debugging, performance validation, and system demonstrations. Once the development phase is complete, this diagnostic data stream can be disabled, allowing the device to operate as a fully autonomous, edge-centric system where no raw visual data leaves the device.

\begin{table*}[!t]
\caption{YOLOv11 comprehensive benchmarks on WOTR, including execution breakdown, memory footprint, and FP32-to-INT8 PTQ accuracy drop. Total \emph{Epochs} are the sum of \emph{NPU}, \emph{HY}, and \emph{SW}. \emph{EC} denotes epoch-controller blobs. $\Delta$ (\%) is computed relative to FP32 $mAP_{50\text{-}95}$ from the non-rounded values. The green-shaded row marks the configurations selected for subsequent experiments.}
\label{tab:yolo_combined_benchmarks}
\centering
\setlength{\tabcolsep}{4.2pt}
\renewcommand{\arraystretch}{1.1}
\resizebox{\textwidth}{!}{%
\begin{tabular}{c c | c c c c c | c c c | c c c}
\hline
\multicolumn{2}{c|}{\textbf{Configuration}} &
\multicolumn{5}{c|}{\textbf{Execution Breakdown}} &
\multicolumn{3}{c|}{\textbf{Memory Footprint}} &
\multicolumn{3}{c}{\textbf{\(mAP_{50\text{--}95}\)}} \\
\textbf{Input} & \textbf{YOLO} &
\textbf{Total} & \textbf{SW} & \textbf{HY} & \textbf{NPU} & \textbf{EC} &
\textbf{Weights [MB]} & \textbf{Act$_{\mathrm{int}}$ [kB]} & \textbf{Act$_{\mathrm{ext}}$ [kB]} &
\textbf{FP32} & \textbf{INT8} & \textbf{$-\Delta\%$} \\
\hline

\multirow{3}{*}{128}
 & n & 155 & 15 & 12 & 128 & 17 & 2.465 & 146  & 0     & 0.16 & 0.14 & 10.01 \\
 & s & 181 & 25 & 18 & 138 & 24 & 8.980 & 292  & 0     & 0.20 & 0.17 & 12.62 \\
 & m & 211 & 25 & 18 & 168 & 24 & 19.112 & 626 & 0     & 0.22 & 0.20 & 12.87 \\
\hline

\multirow{3}{*}{192}
 & n & 172 & 19 & 14 & 139 & 24 & 2.473 & 324  & 0     & 0.22 & 0.19 & 12.56 \\
 & s & 192 & 29 & 18 & 145 & 25 & 8.991 & 673  & 0     & 0.28 & 0.23 & 18.27 \\
 & m & 223 & 29 & 18 & 178 & 25 & 19.126 & 1892 & 576  & 0.32 & 0.26 & 18.69 \\
\hline

\multirow{3}{*}{256}
 & \cellcolor{softgreen!25}\textbf{n}
 & \cellcolor{softgreen!25}\textbf{170}
 & \cellcolor{softgreen!25}\textbf{17}
 & \cellcolor{softgreen!25}\textbf{10}
 & \cellcolor{softgreen!25}\textbf{143}
 & \cellcolor{softgreen!25}\textbf{19}
 & \cellcolor{softgreen!25}\textbf{2.483}
 & \cellcolor{softgreen!25}\textbf{632}
 & \cellcolor{softgreen!25}\textbf{0}
 & \cellcolor{softgreen!25}\textbf{0.28}
 & \cellcolor{softgreen!25}\textbf{0.24}
 & \cellcolor{softgreen!25}\textbf{15.09} \\
 & s & 191 & 29 & 18 & 144 & 25 & 9.004 & 1784 & 512   & 0.35 & 0.28 & 21.02 \\
 & m & 225 & 29 & 18 & 178 & 26 & 19.146 & 3359 & 1792 & 0.40 & 0.31 & 23.88 \\
\hline

\multirow{3}{*}{320}
 & \cellcolor{softgreen!25}\textbf{n}
 & \cellcolor{softgreen!25}\textbf{175}
 & \cellcolor{softgreen!25}\textbf{19}
 & \cellcolor{softgreen!25}\textbf{12}
 & \cellcolor{softgreen!25}\textbf{144}
 & \cellcolor{softgreen!25}\textbf{19}
 & \cellcolor{softgreen!25}\textbf{2.495}
 & \cellcolor{softgreen!25}\textbf{923}
 & \cellcolor{softgreen!25}\textbf{0}
 & \cellcolor{softgreen!25}\textbf{0.33}
 & \cellcolor{softgreen!25}\textbf{0.27}
 & \cellcolor{softgreen!25}\textbf{20.48} \\
 & s & 191 & 29 & 18 & 144 & 25 & 9.018 & 2448 & 800   & 0.41 & 0.31 & 23.02 \\
 & m & 227 & 29 & 18 & 180 & 25 & 19.163 & 4545 & 2804 & 0.46 & 0.34 & 25.24 \\
\hline
\end{tabular}%
}
\end{table*}

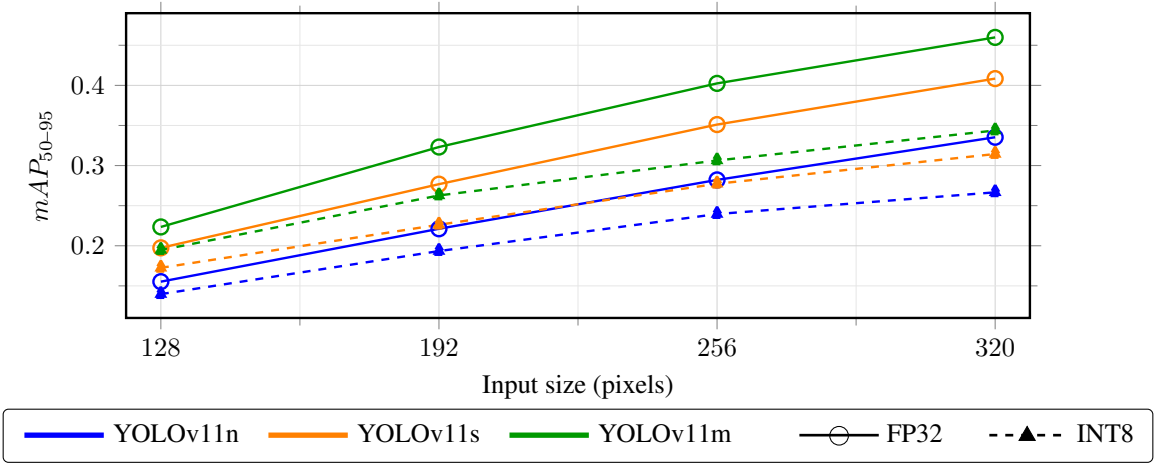
\begin{figure*}[t]
\centering
\begin{tikzpicture}

\begin{axis}[
    name=mainplot,
    width=0.82\textwidth,
    height=0.34\textwidth,
    xlabel={Input size (pixels)},
    ylabel={\(mAP_{50\text{--}95}\)},
    xtick={128,192,256,320},
    xmin=120, xmax=328,
    ymin=0.11, ymax=0.49,
    minor x tick num=1,
    minor y tick num=1,
    grid=both,
    major grid style={line width=.2pt, draw=gray!35},
    minor grid style={line width=.1pt, draw=gray!20},
    tick align=outside,
    mark size=2.8pt,
    line width=0.9pt,
]

\addplot[YOLOv11n, fp32] coordinates { (128,0.155258) (192,0.221185) (256,0.282118) (320,0.335318) };
\addplot[YOLOv11s, fp32] coordinates { (128,0.197407) (192,0.276804) (256,0.351126) (320,0.408434) };
\addplot[YOLOv11m, fp32] coordinates { (128,0.223481) (192,0.323129) (256,0.402469) (320,0.459817) };

\addplot[YOLOv11n, int8] coordinates { (128,0.139720) (192,0.193413) (256,0.239546) (320,0.266652) };
\addplot[YOLOv11s, int8] coordinates { (128,0.172491) (192,0.226228) (256,0.277325) (320,0.314424) };
\addplot[YOLOv11m, int8] coordinates { (128,0.194717) (192,0.262729) (256,0.306368) (320,0.343754) };

\end{axis}

\node[
  anchor=north,
  draw=black,
  line width=0.4pt,
  rounded corners=2pt,
  inner xsep=7pt,
  inner ysep=5pt
] at ($(mainplot.south)+(0,-1.2)$) {%
  \normalsize
  \tikz[baseline=-0.6ex]{\draw[blue, line width=2pt] (0,0)--(1.05,0);}~YOLOv11n
  \hspace{.5em}
  \tikz[baseline=-0.6ex]{\draw[orange, line width=2pt] (0,0)--(1.05,0);}~YOLOv11s
  \hspace{.5em}
  \tikz[baseline=-0.6ex]{\draw[green!60!black, line width=2pt] (0,0)--(1.05,0);}~YOLOv11m
  \hspace{2.0em}
  \tikz[baseline=-0.6ex]{%
    \draw[black, line width=1.0pt] (0,0)--(1.05,0);
    \node at (0.525,0.175) {{\pgfsetplotmarksize{3.5pt}\pgfuseplotmark{o}}};
  }~FP32
  \hspace{1.2em}
  \tikz[baseline=-0.6ex]{%
    \draw[black, dashed, line width=1.0pt] (0,0)--(1.05,0);
    \node at (0.5,0.175) {{\pgfsetplotmarksize{3.5pt}\pgfuseplotmark{triangle*}}};}~INT8
};
\end{tikzpicture}
\caption{{\(mAP_{50\text{--}95}\) versus input size on the WOTR test set for YOLOv11n/s/m in FP32 and INT8 PTQ. INT8 yields consistently lower \(mAP_{50\text{--}95}\) across all model families and input resolutions.}}
\label{fig:map_fp32_int8}
\end{figure*}

\section{Results and Discussion}
\label{sec:Results}

Our experimental evaluation begins with a deployment-oriented analysis to select a YOLOv11 configuration optimized for the ARGO platform. We evaluate this configuration based on accuracy, memory requirements, and hardware performance. Following this, we report system-level measurements, detailing the power profiles of the main circuit components during active execution.

\subsection{Model Selection for On-Device Deployment}

Following the training and evaluation campaign described in Section~\ref{sec:modello}, we benchmarked the complete grid of YOLOv11 variants (n/s/m) and input sizes (128, 192, 256, 320) on the WOTR test set.
For each configuration, we report not only detection accuracy but also the deployment-relevant characteristics on ARGO, namely the weight footprint, the allocation of intermediate activations between internal and external memory, and the execution breakdown across software (SW), hybrid (HY), and NPU execution epochs.

Following the preliminary evaluation, a single optimized configuration was selected for deployment in the final firmware to undergo comprehensive on-device characterization.

The deployed configuration is selected by enforcing the following deployment-oriented requirements:
(i) \textbf{NPU-centric execution} (predominantly NPU epochs, with limited HY/SW partitions);
(ii) \textbf{internal-only activations} (no external activation allocation, i.e., Act$_{\mathrm{ext}}=0$); and
(iii) \textbf{compact weight footprint} (i.e., within the non-volatile budget).

Table~\ref{tab:yolo_combined_benchmarks} reports FP32 vs INT8 \(mAP_{50\text{--}95}\) across all trained configurations, and it reports the corresponding memory footprints and the execution breakdown on the target platform.
Fig.~\ref{fig:map_fp32_int8} complements this table by visualizing the FP32 versus INT8 trends across model families and input sizes.

Within the explored grid, YOLOv11n at \(256\times256\) and \(320\times320\) emerge as the two configurations that satisfy all deployment-oriented constraints.
The \(256\times256\) configuration is selected as the final operating point, as it ensures fully internal activation allocation with lower volatile memory pressure, while maintaining competitive accuracy. Furthermore, with this input size, the network latency is significantly reduced, achieving a 28\% reduction compared to the \(320\times320\) configuration, guaranteeing the execution of the complete pipeline at 10 Hz with a system clock frequency of 600 MHz.
The selected configuration is characterized by 143 NPU epochs out of 170 total (10 HY and 17 SW epochs, with 19 epoch-controller blobs), Act$_{\mathrm{ext}}=0$ with Act$_{\mathrm{int}}=632\,\mathrm{kB}$, and a weight footprint of 2.483\,MB.
In terms of accuracy, it achieves \(mAP_{50\text{--}95}=0.28\) in FP32 and \(mAP_{50\text{--}95}=0.24\) after INT8 PTQ ($\%\Delta=-15.09\%$).
Accordingly, we deploy YOLOv11n at $256\times256$ input size.

Numerical agreement between the host-side reference execution and the on-device execution on the STM32N6 was evaluated over the full WOTR test set ($N=2534$).
The global cosine similarity and the global Pearson correlation were both equal to $0.9996$.
The global relative $\ell_2$ error was $0.028$, while the normalized root mean square error (NRMSE) was $0.030$.

\begin{figure}[t]
    \centering

    \begin{subfigure}[t]{0.485\columnwidth}
        \captionsetup{font=footnotesize}
        \centering
        \includegraphics[width=0.9\linewidth]{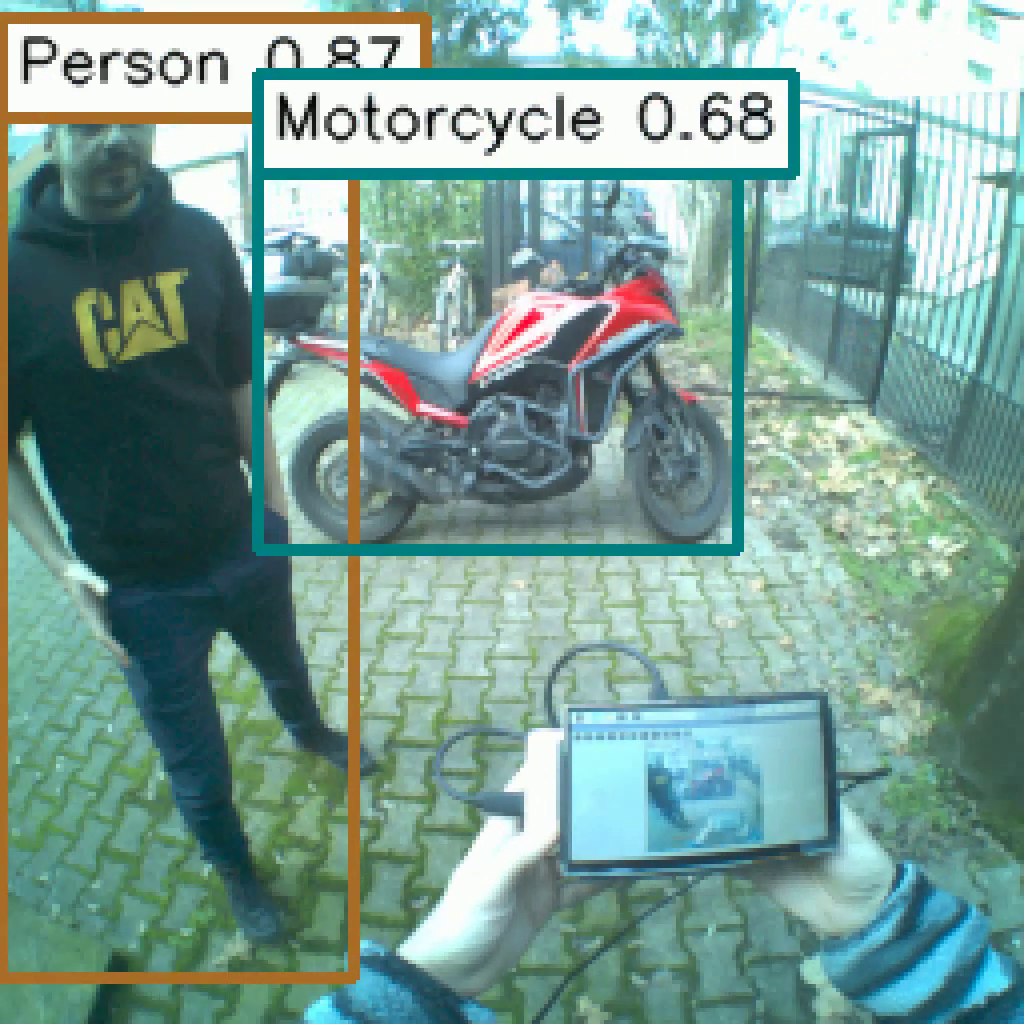}
        \caption*{\texttt{A.}}
    \end{subfigure}\hfill
    \begin{subfigure}[t]{0.485\columnwidth}
        \captionsetup{font=footnotesize}
        \centering
        \includegraphics[width=0.9\linewidth]{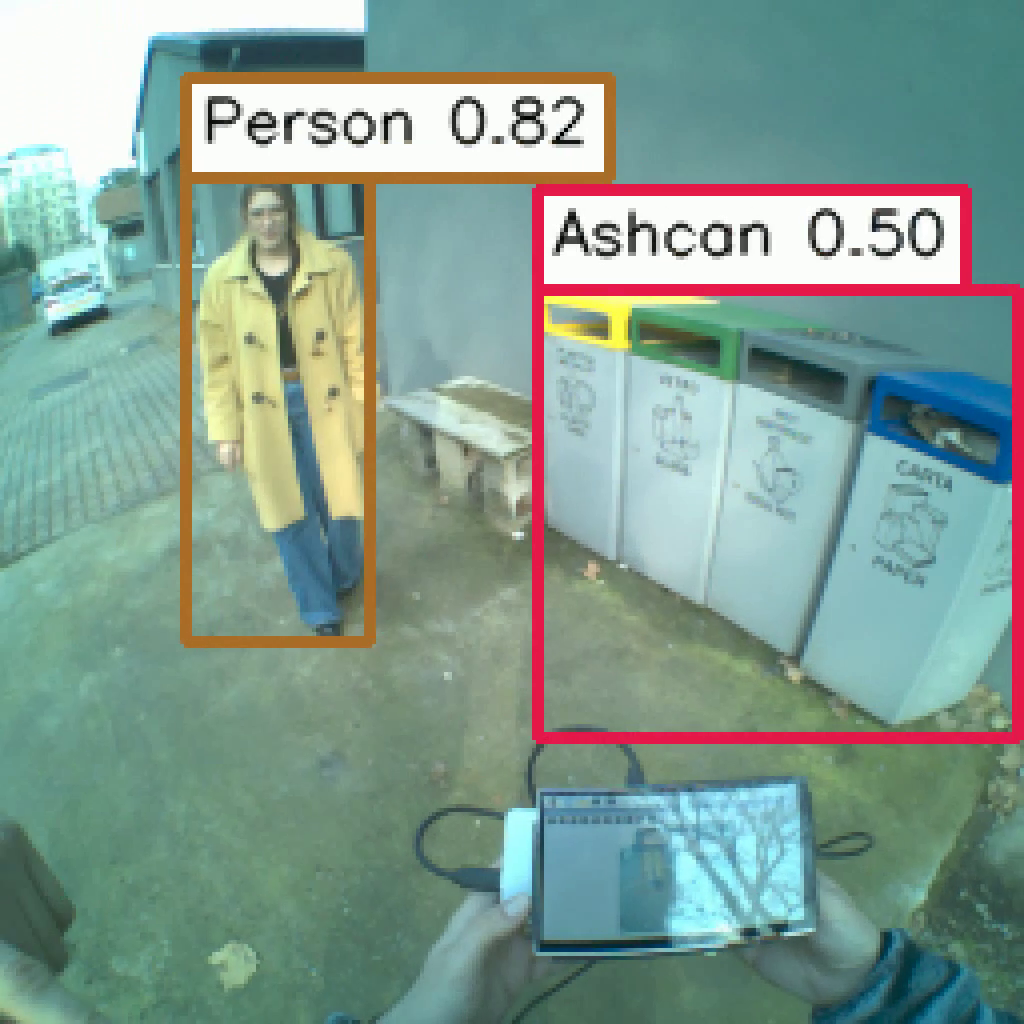}
        \caption*{\texttt{B.}}
    \end{subfigure}

    \vspace{0.6em}

    \begin{subfigure}[t]{0.485\columnwidth}
        \captionsetup{font=footnotesize}
        \centering
        \includegraphics[width=0.9\linewidth]{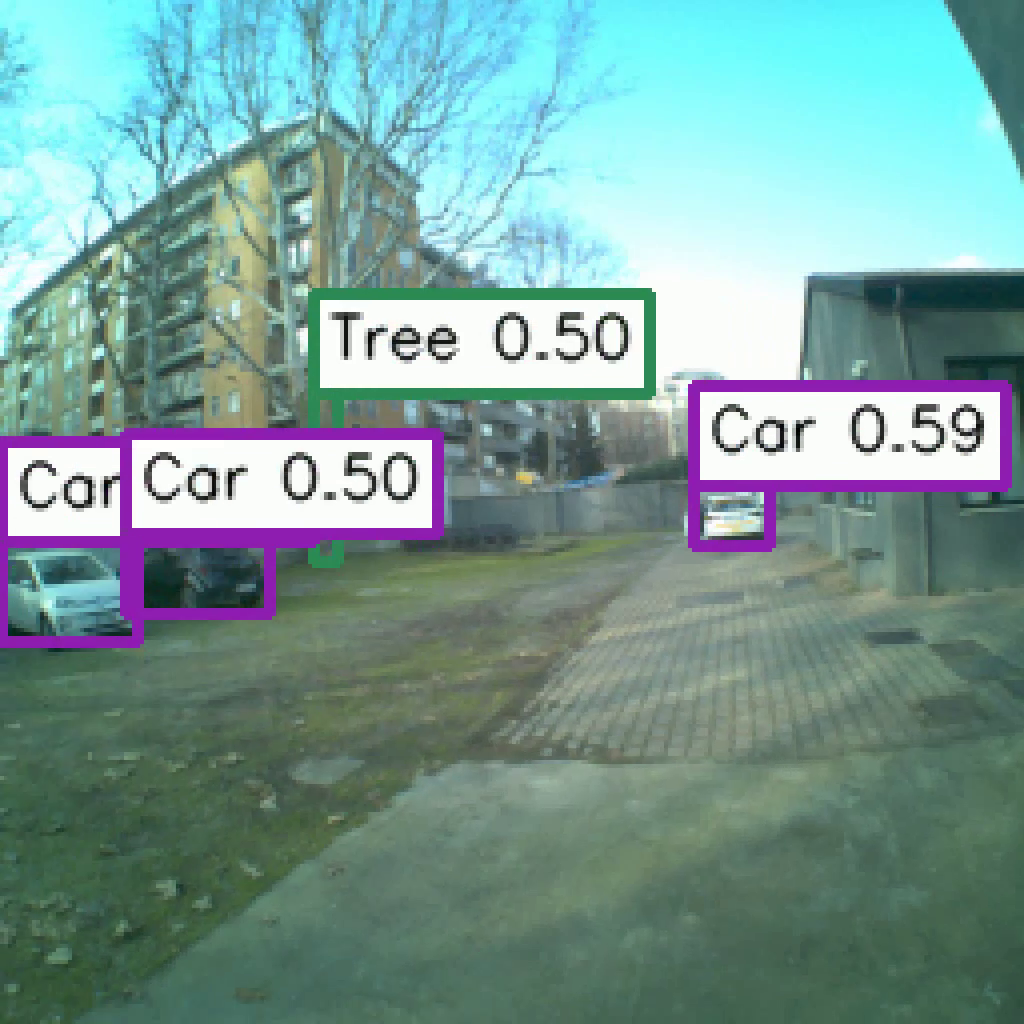}
        \caption*{\texttt{C.}}
    \end{subfigure}

    \caption{Qualitative on-device detection examples produced by ARGO using the deployed YOLOv11n model. Predicted bounding boxes are shown with class labels and confidence scores after the embedded post-processing; detections are overlaid on the corresponding RGB frames for visualization.}
    \label{fig:ARGO_example}
\end{figure}

Fig.~\ref{fig:ARGO_example} shows representative qualitative detections obtained on-device with the deployed YOLOv11n configuration, highlighting typical urban obstacles and infrastructure elements as seen from the egocentric viewpoint of the eyewear.

Taken together, these results provide a system-level view of the trade-offs across the explored design space, highlighting a consistent relationship between accuracy, memory, and execution cost.

As the input size increases, detection accuracy improves, but the activation footprint also grows, eventually requiring external memory allocation for the larger configurations.
In an embedded setting, this aspect is particularly critical, since relying on external memory implies activating additional hardware resources, which increases system-level energy consumption.

A second trend concerns quantization.
INT8 PTQ consistently reduces \(mAP_{50\text{--}95}\) with respect to FP32, and the degradation becomes more pronounced for larger models, reaching up to \(25.24\%\) for YOLOv11m at \(320\times320\).
This suggests that, within the explored design space, increased model capacity improves absolute accuracy at the cost of higher sensitivity to PTQ.

More generally, the measured \(mAP_{50\text{--}95}\) values remain below those typically associated with standard YOLO deployments, since the models are trained and deployed at input resolutions substantially smaller than the default \(640\times640\). On the target platform, however, larger inputs would increase both the computational load and the camera active time, thereby raising latency and overall energy consumption.

In this sense, higher accuracy is not obtained for free: improving detection performance requires either larger inputs or larger models, both of which demand additional memory, compute, and energy.

These results highlight that, in an embedded setting, accuracy must be interpreted jointly with system-level constraints rather than as an isolated metric.
Accordingly, the selected configuration represents an operating point that maximizes detection performance while ensuring fully on-device execution, bounded memory usage, and energy-efficient operation.

The execution breakdown further supports this interpretation.
The most deployment-relevant configurations remain strongly NPU-centered, with the large majority of epochs mapped to the hardware accelerator and only a limited fraction assigned to hybrid or software execution.
For instance, YOLOv11n at \(256\times256\) executes 143 out of 170 epochs on the NPU, while YOLOv11n at \(320\times320\) executes 144 out of 175 epochs on the NPU. This is an important property for embedded deployment, since a higher share of HY/SW execution would introduce additional overhead on the general-purpose processing path, reducing both execution efficiency and system-level energy effectiveness.

Finally, the weight footprint behaves as expected.
For a given model family, it is primarily determined by the model scale itself, with only a minor overhead across input resolutions.
In contrast, volatile memory and execution cost are much more sensitive to the input size, making the activation footprint the dominant constraint in practical on-device deployment.

From a numerical standpoint, the agreement reported above indicates that the deployed implementation faithfully reproduces the host-side reference behavior.
Therefore, the observed system-level trade-offs can be attributed to the deployment constraints themselves rather than to relevant output distortions introduced by compilation or execution on the target platform.

\subsection{Power and Energy Characterization}

To quantify the effectiveness of the proposed co-design, we performed a comprehensive energy characterization of the ARGO platform during real-time inference. This analysis evaluates the power profile of each circuit block to assess the system's overall sustainability and autonomy.

The analysis focused on the main power rails of the system: the N6 MCU, which manages the neural model execution; the three power rails of the image sensor; and the external FLASH memory. To measure power consumption without altering the system's operating point, we adopted the benchmarking methodology described in~\cite{11228997}. Shunt resistors were inserted in series with the power rails, with resistance values selected to ensure a negligible voltage drop while maintaining a measurable signal-to-noise ratio for oscilloscope acquisition.

\begin{figure}[!t]
    \centering
    \includegraphics[width=\linewidth]{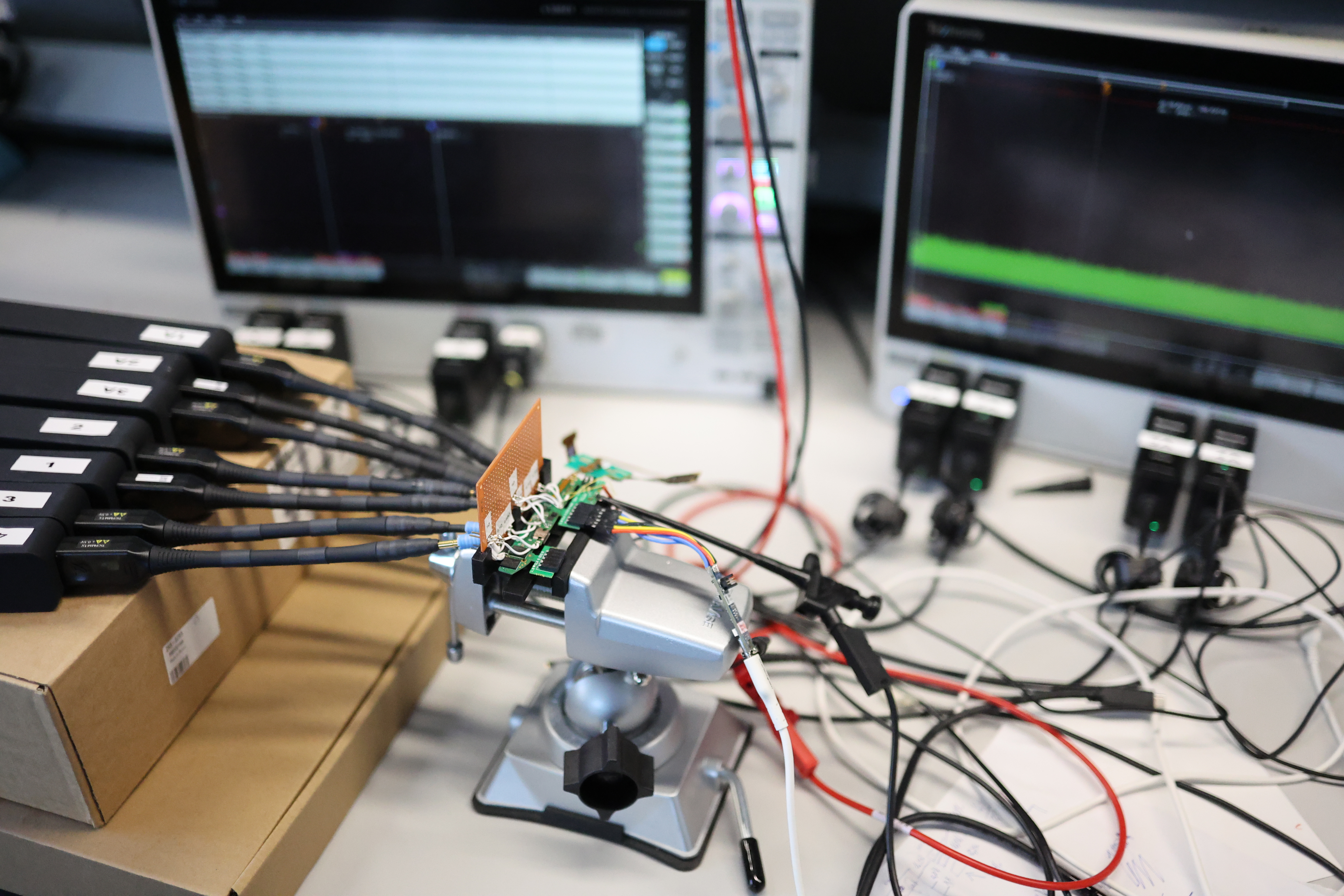}
    \caption{Experimental measurement setup. Two synchronized oscilloscopes with a common trigger sample the shunt resistor voltages via differential probes during application execution.}
    \label{fig:foto_setup_misura}
\end{figure}

The experimental setup, shown in Fig.~\ref{fig:foto_setup_misura}, consists of two synchronized oscilloscopes sharing a common trigger. Differential probes were employed to measure the voltage drop across the shunt resistors, while one single-ended probe was used to capture the FSM hardware triggers, enabling the temporal windowing of each operational phase.

\begin{figure*}[t]
    \centering
    \includegraphics[width=.9\linewidth]{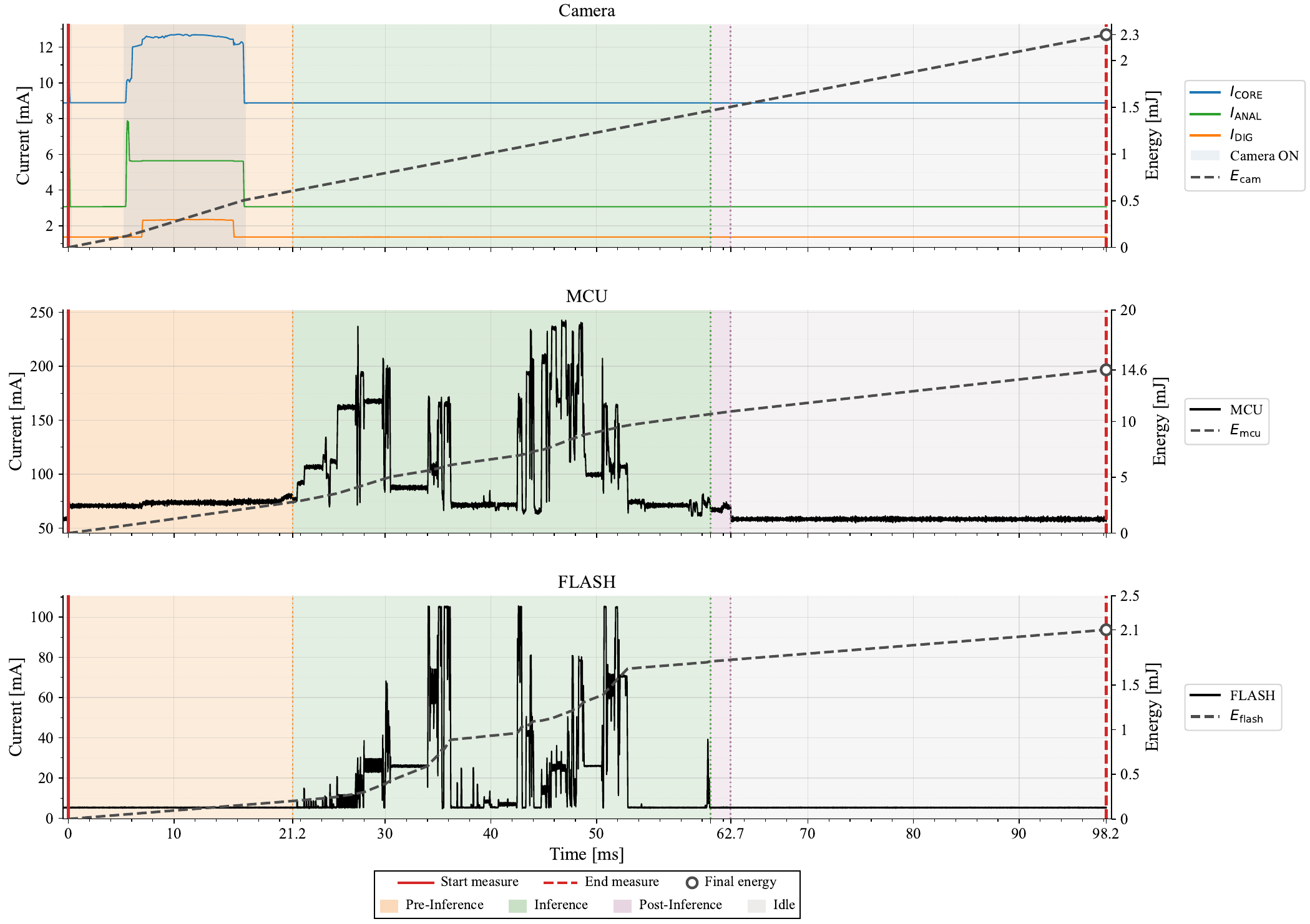}
    \caption{Power consumption breakdown of the main circuit blocks during one application cycle. Red vertical lines mark the start and end of the FSM cycle, while dashed curves report the cumulative energy per cycle. Current traces are displayed after moving-average filtering for visualization only. The top, middle, and bottom panels show the image sensor, STM32N6 microcontroller, and external FLASH memory, respectively.}
    \label{fig:risultati_potenza}
\end{figure*}

The measurement results are illustrated in Fig.~\ref{fig:risultati_potenza}.

In this configuration, the N6 MCU operates at sub-maximal clock frequencies: specifically 600 MHz for the CPU, 600 MHz for the NPU, and 600 MHz for the internal RAM. This frequency scaling strategy ensures a sufficient frame rate (FPS) for the target application while minimizing power consumption. The efficiency of this approach is rooted in the dynamic power consumption of CMOS circuits. To achieve higher clock frequencies, the MCU must increase its internal supply voltage ($V_{core}$), which impacts power dissipation quadratically according to the following relation:$$P \approx C \cdot V_{dd}^2 \cdot f$$ where $P$: Total dynamic power dissipation. $C$: Total equivalent capacitance of the clock distribution network. $V_{core}$: Operating supply voltage. $f$: Switching frequency of the clock signal. Consequently, by capping the frequency at 600 MHz, the system can operate at a lower voltage level, leading to a significant reduction in power dissipation.

The vertical red line in Fig.~\ref{fig:risultati_potenza} denote the application's start-of-cycle. Initially, the system performs internal register updates and commands the image sensor to transition from standby to active acquisition. After the frame is captured and transmitted, the sensor reverts to its standby, yet not low power, state, resulting in an energy consumption of 2.3 mJ per cycle. Subsequently, the MCU utilizes the DCMIPP peripheral to execute ISP algorithms. Following this, the image undergoes a channel transposition (from interleaved to planar mode) to meet the specific input requirements of the neural network. Finally, the resized frame is transferred to the NPU for inference.

During the inference phase, the MCU retrieves model weights from the external FLASH memory. This stage represents the system's peak energy consumption.
In contrast, the post-processing phase is characterized by a very short duration and negligible power overhead. Since the entire processing pipeline completes in less than 100 ms, the N6 MCU enters a light low-power state for the remainder of the cycle. It should be noted that no aggressive power-management strategies were implemented on the microcontroller during these tests; therefore, the energy consumption during the acquisition phase remains subject to further optimization.
These results therefore represent a worst-case power consumption scenario.

\begin{figure}[!htpb]
    \centering
    \includegraphics[width=\linewidth]{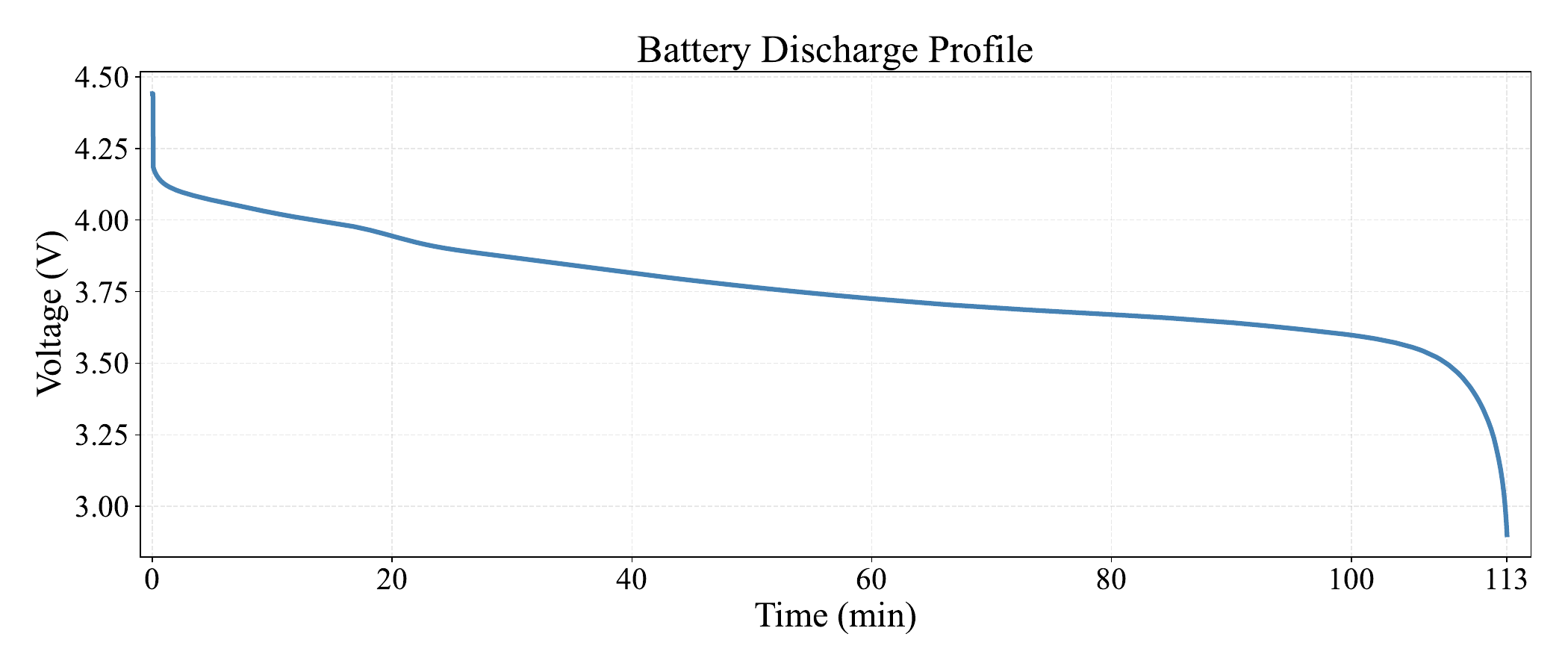}
    \caption{Full battery discharge profile obtained with a digital multimeter. The measurement begins upon USB cable removal. The battery voltage discharges from a full charge of 4.2V, stabilizing at the typical 3.7V plateau before rapidly dropping to the 2.9V cut-off threshold enforced by the PMIC and BMS for battery protection.}
    \label{fig:profilo-scarica}
\end{figure}

To complete the experimental evaluation, we measured the full battery discharge profile of the system when powered by a 200~mAh Li-Po battery---a capacity specifically chosen to meet the ergonomic requirements of smart eyewear. Fig.~\ref{fig:profilo-scarica} illustrates the characteristic discharge curve as the system operates.
The measurements were captured using a digital multimeter connected to the battery's positive test point and ground.
The experimental procedure began with the PMIC fully charging the battery while the system was active and powered via a 5V USB connection. Upon disconnecting the cable, the PMIC automatically transitioned the power source from the USB bus to the battery. The system then remained operational until reaching the $2.9V$ cut-off voltage, a threshold enforced by both the PMIC and the integrated Battery Management System (BMS) of the battery pack. The total system autonomy was derived directly from the measurement timescale, and is approximately 113 minutes. This empirical value shows a slight deviation from theoretical estimates calculated by summing the individual contributions of the MCU, camera, flash memory, and BLE module. This discrepancy is attributed to secondary power sinks, including oscillators, resistor dividers, leakage currents, and conversion losses from both DC-DC converters and LDO regulators. Nonetheless, the results confirm that the MCU remains the primary contributor to power consumption, followed by the camera module.

\begin{table*}[t]
\caption{MCU and FLASH rail current statistics over $N=1000$ acquisitions. The \textit{MAIN} rail feeds the on-chip N6 regulator generating $V_{\mathrm{core}}$, while the \textit{LDO} rail supplies analog, IO, CSI, and USB peripherals. Both rails operate at $1.8\,\mathrm{V}$.}
\label{tab:mcu_rails_1k}
\centering
\setlength{\tabcolsep}{3.5pt}
\renewcommand{\arraystretch}{1.05}
\resizebox{\textwidth}{!}{%
\begin{tabular}{c|cccc|cccc|cccc|cccc}
\hline
 & \multicolumn{4}{c|}{$I_{\mathrm{MAIN}}$ [mA]}
 & \multicolumn{4}{c|}{$I_{\mathrm{LDO}}$ [mA]}
 & \multicolumn{4}{c|}{$I_{\mathrm{TOT}}$ [mA]}
 & \multicolumn{4}{c|}{$I_{\mathrm{FLASH}}$ [mA]}\\
\hline
& \textbf{Mean$\pm$Std} & \textbf{Min} & \textbf{Max} & \textbf{Pk--Pk}
& \textbf{Mean$\pm$Std} & \textbf{Min} & \textbf{Max} & \textbf{Pk--Pk}
& \textbf{Mean$\pm$Std} & \textbf{Min} & \textbf{Max} & \textbf{Pk--Pk}
& \textbf{Mean$\pm$Std} & \textbf{Min} & \textbf{Max} & \textbf{Pk--Pk}  \\
\hline
Pre-inference
& $71.56 \pm 0.12$& 71.27& 71.77& 0.50& $0.8 \pm 0.01$& 0.77& 0.9& 0.1& $72.36 \pm 0.12$& 72.04& 72.67& 0.60& $5.21\pm 0.07$& 4.93& 5.29&0.36\\

Inference
& $109.61 \pm 0.11$& 109.36& 109.78& 0.41& $0.96 \pm 0.01$& 0.94& 1.0& 0.1& $110.57 \pm 0.11$& 110.30& 110.78& 0.51& $21.9\pm 0.076$& 21.66& 21.98&0.31\\

Post-inference
& $67.26 \pm 0.10$& 66.93& 67.47& 0.544& $0.79 \pm 0.01$& 0.76& 0.88& 0.1
& $68.05 \pm 0.10$& 67.69& 68.35& 0.64& $5.21\pm 0.074$& 4.92& 5.29&0.36\\

Idle
& $57.52\pm 0.25$& 56.70& 58.05& 1.34& $0.79\pm 0.01$& 0.76& 0.87& 0.11& $58.31\pm 0.25$& 57.46& 58.92& 1.45& $5.19\pm 0.07$& 4.93& 5.27&0.34\\
\hline
\end{tabular}}
\end{table*}

\begin{table}[t]
\caption{Current statistics over $N=1000$ acquisitions for the three camera rails. The values are relatives to the pre-inference phase only.}
\label{tab:meas_contrib_1k}
\centering
\setlength{\tabcolsep}{3.5pt}
\renewcommand{\arraystretch}{1.05}
\resizebox{0.95\linewidth}{!}{%
\begin{tabular}{l|c|c|c|c}
\hline
 & \textbf{Mean$\pm$Std [mA]} & \textbf{Min [mA]} & \textbf{Max [mA]} & \textbf{Pk--Pk [mA]} \\
\hline
$I_{CoreCam}$   & $10.42 \pm 3.21$ & 10.40   & 10.426  & 0.017  \\
$I_{DigCam}$   & $1.66  \pm 0.01$ & 1.64  & 1.677   & 0.032   \\
$I_{AnCam}$   & $4.14 \pm 0.004$ & 4.14   & 4.15  & 0.011  \\
\hline
\end{tabular}}
\end{table}

\section{Conclusion and Future Development}
\label{sec:Future development}
The ARGO platform represents a step toward smart eyewear capable of standalone, on-device execution of complex neural models. By operating independently of cloud connectivity, the system ensures that sensitive data remains local, inherently guaranteeing a high level of user privacy. Furthermore, its flexible hardware architecture enables the acquisition of synchronized, multimodal datasets from an egocentric perspective, providing a valuable tool for the development and refinement of future AI models.

Although the current system is highly optimized, further energy savings could be achieved by implementing aggressive low-power strategies. Future iterations will explore fine-grained power gating on unused voltage rails and advanced microcontroller sleep states to further minimize the energy footprint. Conversely, for applications requiring higher temporal resolution, the frame rate could be increased by boosting the operational frequency and implementing a more efficient pipeline. Specifically, parallelizing camera frame acquisition with NPU execution would significantly enhance throughput, albeit at the cost of higher power consumption.

Ultimately, this work highlights the necessity of a multidisciplinary approach in developing complex embedded systems. The convergence of hardware design, firmware optimization, and machine learning is essential to reach the performance frontier. The evolution of embedded processors and neural accelerators must proceed in lockstep with algorithmic innovation to eventually enable the seamless execution of sophisticated AI entirely at the edge, eliminating the latency, cost, and privacy risks associated with persistent server connectivity.

\section*{Acknowledgments}
\textit{This work was carried out in the EssilorLuxottica "Smart Eyewear Lab", a Joint Research Center between EssilorLuxottica and Politecnico di Milano.
}

The authors would like to thank Fabio D'Arcangelo, Gianluca Candiotto, and Dino Michelon from the NTI team at Luxottica (Agordo, Italy) for their technical support and expertise in the design of the plastic frame.

Special thanks are also due to Efrem Festini Cappello from the Smart Eyewear Lab (Milan, Italy) for his assistance in managing the 3D printing process.

Special thanks also go to David Siorpaes and Alessandro Vaghi, both currently at STMicroelectronics, for their technical support with the N6 microcontroller.

Portions of this manuscript were revised for language clarity and syntactic accuracy with the assistance of Claude, an AI language model developed by Anthropic.

\bibliographystyle{unsrt}    
\bibliography{bibliography}

\end{document}